\documentclass[10pt,twocolumn,letterpaper]{article}

\usepackage[pagenumbers]{cvpr}

\usepackage{times}
\usepackage{epsfig}
\usepackage{graphicx}
\usepackage{amsmath}
\usepackage{amssymb}
\usepackage{xcolor}
\usepackage{booktabs}
\usepackage{cite}
\usepackage{microtype}
\usepackage{enumitem}
\usepackage[accsupp]{axessibility}

\definecolor{cvprblue}{rgb}{0.21,0.49,0.74}
\usepackage[breaklinks,colorlinks,citecolor=cvprblue]{hyperref}

\setitemize{noitemsep,topsep=0pt,parsep=0pt,partopsep=0pt}

\newcommand{\red}[1]{\textcolor{red}{#1}}
\newcommand{\blue}[1]{\textcolor{blue}{#1}}

\def\VspaceL{\vspace{-0.20cm}}
\def\VspaceS{\vspace{-0.10cm}}
\def\VspaceSF{\vspace{-0.10cm}}
\def\VspaceST{\vspace{-0.00cm}}

\def\VspaceSUPP{\vspace{-0.45cm}}
\def\VspaceSUPPS{\vspace{-0.30cm}}

\def\confName{CVPR}
\def\confYear{2024}

\title{Hourglass Tokenizer for Efficient Transformer-Based 3D Human Pose Estimation}

\author{
Wenhao Li\textsuperscript{1} \quad
Mengyuan Liu\textsuperscript{1}\thanks{Corresponding Author.} \quad
Hong Liu\textsuperscript{1} \quad
Pichao Wang\textsuperscript{2}\thanks{The work does not relate to author's position at Amazon.} \quad
Jialun Cai\textsuperscript{1} \quad
Nicu Sebe\textsuperscript{3}
\\
\textsuperscript{1}{\normalsize National Key Laboratory of General Artificial Intelligence, Peking University, Shenzhen Graduate School} \\
\textsuperscript{2}{\normalsize Amazon Prime Video} \quad
\textsuperscript{3}{\normalsize University of Trento} \\
{\tt\small \{wenhaoli,liumengyuan,hongliu\}@pku.edu.cn}  \quad{\tt\small pichaowang@gmail.com} \\  {\tt\small cjl@stu.pku.edu.cn} \quad {\tt\small niculae.sebe@unitn.it}
}

\begin{document}
\maketitle

\begin{abstract}
Transformers have been successfully applied in the field of video-based 3D human pose estimation. However, the high computational costs of these video pose transformers (VPTs) make them impractical on resource-constrained devices. In this paper, we present a plug-and-play pruning-and-recovering framework, called Hourglass Tokenizer (HoT), for efficient transformer-based 3D human pose estimation from videos. Our HoT begins with pruning pose tokens of redundant frames and ends with recovering full-length tokens, resulting in a few pose tokens in the intermediate transformer blocks and thus improving the model efficiency. To effectively achieve this, we propose a token pruning cluster (TPC) that dynamically selects a few representative tokens with high semantic diversity while eliminating the redundancy of video frames. In addition, we develop a token recovering attention (TRA) to restore the detailed spatio-temporal information based on the selected tokens, thereby expanding the network output to the original full-length temporal resolution for fast inference. Extensive experiments on two benchmark datasets (i.e., Human3.6M and MPI-INF-3DHP) demonstrate that our method can achieve both high efficiency and estimation accuracy compared to the original VPT models. For instance, applying to MotionBERT and MixSTE on Human3.6M, our HoT can save nearly 50\% FLOPs without sacrificing accuracy and nearly 40\% FLOPs with only 0.2\% accuracy drop, respectively. Code and models are available at \url{https://github.com/NationalGAILab/HoT}. 
\vspace{-0.6cm}
\end{abstract}

\section{Introduction}
3D human pose estimation (HPE) from videos has numerous applications, such as action recognition \cite{liu2017enhanced,wang2018depth,luvizon2020multi}, human-robot interaction \cite{zimmermann20183d,garcia2019human}, and computer animation \cite{mehta2017vnect}. 
Current video-based 3D HPE methods mainly follow the pipeline of 2D-to-3D pose lifting \cite{stgcn,zeng2020srnet,hu2021conditional,zhang2022uncertainty,you2023co,chen2023hdformer,liu2023posynda}. 
This two-stage pipeline first utilizes an off-the-shelf 2D HPE model to detect 2D body joints for each video frame and then employs a separate lifting model to estimate 3D pose sequences from the detected 2D poses. 

Recently, transformer-based architectures \cite{poseformer,mhformer,mixste,motionbert} have shown state-of-the-art (SOTA) performance in the field of video-based 3D HPE, since they are effective at modeling the long-range dependencies among video frames. 
These video pose transformers (VPTs) typically regard each video frame as a pose token and utilize extremely long video sequences to achieve superior performance (\textit{e.g.}, 81 frames in \cite{poseformer}, 243 frames in \cite{mixste,pstmo,motionbert}, or 351 frames in \cite{stride,mhformer,einfalt2023uplift}). 
However, these methods inevitably suffer from high computational demands since the VPT's self-attention complexity grows quadratically with respect to the number of tokens (\textit{i.e.}, frames), hindering the deployment of these heavy VPTs on devices with limited computing resources. 

\begin{figure}[t]
\centering
\includegraphics[width=1.00\linewidth]{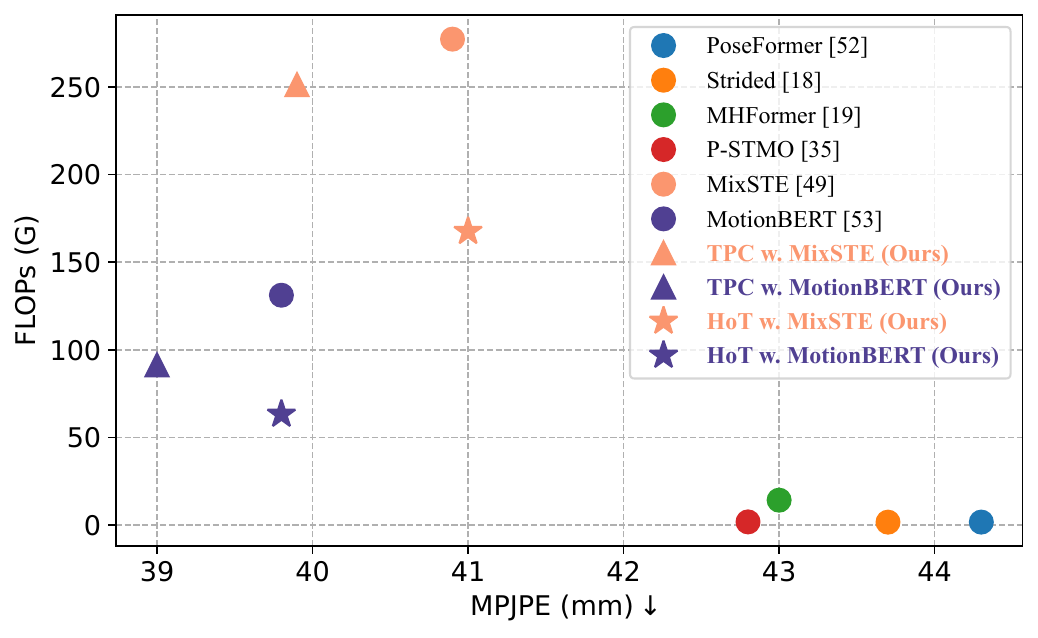}
\VspaceSF
\caption
{
FLOPs and estimation errors (MPJPE, lower is better) of different VPTs on Human3.6M dataset. 
We achieve highly competitive or even better results while saving FLOPs. 
}
\label{fig:flops}
\VspaceL
\end{figure}

To achieve efficient VPTs, two crucial factors require careful consideration:
\textbf{(i)}
Directly reducing the frame number can boost VPTs' efficiency, but it results in a small temporal receptive field that limits the model to capture richer spatio-temporal information to improve performance \cite{videopose,liu2020attention}. 
Hence, it is essential to design an efficient solution while maintaining a large temporal receptive field for accurate estimation. 
\textbf{(ii)}
Adjacent frames in a video sequence contain redundant information due to the similarity of nearby poses (50 Hz cameras used in Human3.6M \cite{ionescu2013human3}). 
Moreover, recent studies \cite{rao2021dynamicvit,wang2022vtc,li2022sait} found that some tokens tend to be similar in the deep transformer blocks. 
Thus, we infer that using full-length pose tokens in these blocks leads to redundant calculations and contributes little to the final estimation. 

Based on these observations, we propose to prune pose tokens in the deep transformer blocks to improve the efficiency of VPTs. 
Although token pruning can reduce the number of tokens and bring efficiency, it also makes it difficult to estimate the consecutive 3D pose of all frames, as in existing VPTs \cite{mhformer,mixste,motionbert}, where each token corresponds to a frame. 
Additionally, for efficient inference, a real-world 3D HPE system should be able to estimate the 3D poses of all frames at once in an input video. 
Therefore, in order to make our method more compatible with being plugged into existing VPTs and achieve fast inference, we need to recover the original full-length tokens for all-frame estimation. 

\begin{figure}[t]
\centering
\includegraphics[width=1.00\linewidth]{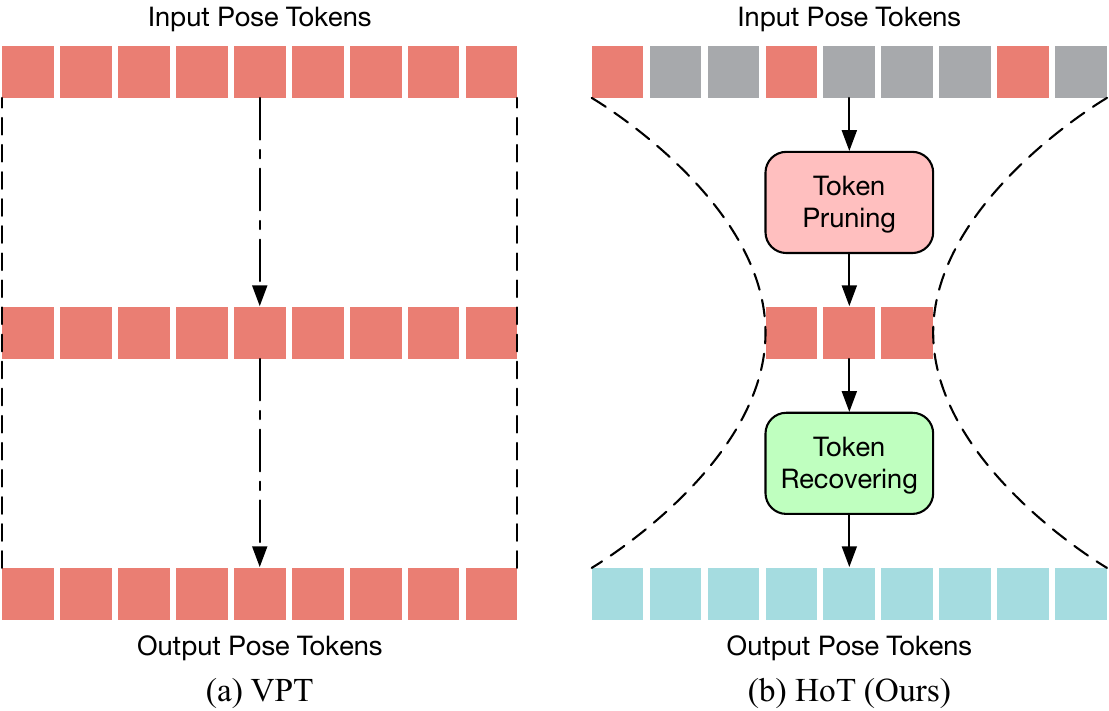}
\VspaceSF
\caption
{
\textbf{(a)} Existing VPTs follow a ``rectangle'' paradigm that retains the full-length sequence across all blocks, which incurs expensive and redundant computational costs. 
\textbf{(b)} Instead, our HoT follows an ``hourglass'' paradigm that prunes the pose tokens and recovers the full-length tokens, which keeps a few tokens in the intermediate transformer blocks and thus improves the model efficiency. 
The gray squares represent the pruned tokens. 
}
\label{fig:pipeline}
\VspaceL
\end{figure}

Driven by this analysis, we present a novel \textit{pruning-and-recovering} framework for efficient transformer-based 3D HPE from videos. 
Different from existing VPTs that maintain the full-length sequence across all blocks, our method begins with pruning the pose tokens of redundant frames and ends with recovering the full-length tokens. 
By using these two designs, we can keep only a few tokens in the intermediate transformer blocks and thus improve the model efficiency (see Figure~\ref{fig:pipeline}). 
For this to be achieved effectively, we argue that the key is to select a few representative tokens with high semantic diversity, as such tokens can maintain rich information while reducing video redundancy. 
Since the cluster centers can retain the semantic diversity of the original signal, we propose a token pruning cluster (TPC) module that utilizes the cluster to dynamically select the cluster centers as the representative tokens. 
Furthermore, we develop a lightweight token recovering attention (TRA) module to restore the detailed spatio-temporal information based on the selected tokens, which expands the low temporal resolution caused by pruning operation to the full temporal resolution. 
This strategy enables the network to estimate consecutive 3D poses of all frames, which facilitates fast inference. 

\begin{figure*}[tb]
\centering
\includegraphics[width=1.00\linewidth]{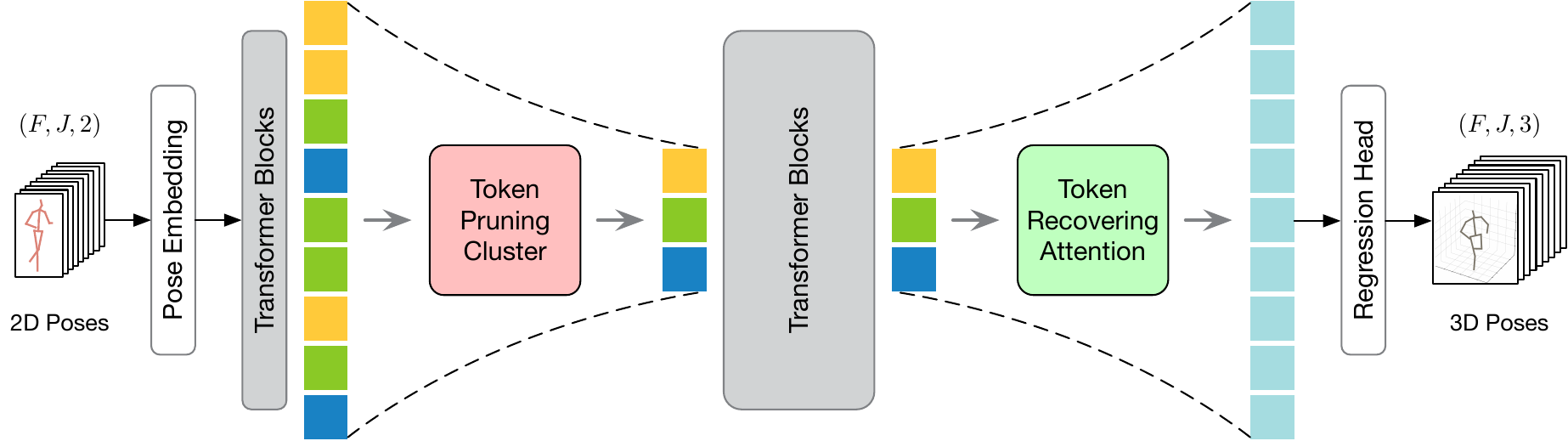}
\VspaceSF
\caption
{
Overview of the proposed Hourglass Tokenizer (HoT).
It mainly consists of a token pruning cluster (TPC) module and a token recovering attention (TRA) module. 
TPC selects the pose tokens of representative frames after the first few transformer blocks and TRA recovers the full-length tokens after the last transformer block. 
}
\label{fig:overview}
\VspaceL
\end{figure*}

Our method can be easily integrated into existing VPTs \cite{mhformer,mixste,motionbert} with minimal modifications (see Figure~\ref{fig:overview}). 
Specifically, the first few transformer blocks of VPTs remain unchanged to obtain pose tokens with comprehensive information from full video frames. 
These pose tokens are then pruned by our TPC, and the remaining tokens that serve as the representative tokens are further fed into the subsequent transformer blocks. 
Finally, the full-length tokens are recovered by TRA, which is added after the last transformer block, while the intermediate transformer blocks still use representative tokens. 
Thus the additional parameters and FLOPs from TRA are negligible. 
Since the number of tokens first decreases through pruning and then increases through recovering, we refer to the framework as an hourglass \cite{newell2016stacked} and name it as Hourglass Tokenizer (HoT). 

To validate the effectiveness and efficiency of our method, we deploy it on top of SOTA VPTs (MHFormer \cite{mhformer}, MixSTE \cite{mixste}, and MotionBERT \cite{motionbert}). 
Extensive experiments demonstrate that existing VPTs consume huge unnecessary computational costs in capturing temporal information, and the proposed method can not only maintain the ability of the model but also reduce the computational costs. 
As shown in Figure~\ref{fig:flops}, our HoT can reduce nearly 50\% floating-point operations (FLOPs) on MotionBERT \cite{motionbert} without sacrificing performance and nearly 40\% FLOPs on MixSTE \cite{mixste} with only 0.2\% performance loss. 

The contributions of our paper are summarized below:
\begin{itemize}
\item We present HoT, a plug-and-play \textit{pruning-and-recovering} framework for efficient transformer-based 3D HPE from videos. 
Our HoT reveals that maintaining the full-length pose sequence is redundant, and a few pose tokens of representative frames can achieve both high efficiency and performance. 
\item To accelerate VPTs effectively, we propose a TPC module to select a few representative tokens for video redundancy reduction and a TRA module to restore the original temporal resolution for fast inference. 
\item Extensive experiments conducted on three recent VPTs show that HoT achieves highly competitive or even superior results while significantly improving efficiency. 
\end{itemize}

\section{Related Work} 
\noindent \textbf{Transformer-based 3D HPE.}
Transformers are firstly proposed in \cite{transformer} and have been successfully applied to video-based 3D HPE \cite{poseformer,mhformer,mixste,motionbert}. 
These video pose transformers (VPTs) are often built to capture spatial and temporal information for 3D HPE using transformers. 
For instance, 
MHFormer \cite{mhformer} learns spatio-temporal multi-hypothesis representations of 3D human poses via transformers. 
MixSTE \cite{mixste} proposes a mixed spatio-temporal transformer to capture the temporal motion of different body joints. 
MotionBERT \cite{motionbert} presents a dual-stream spatio-temporal transformer to model long-range spatio-temporal relationships among skeletal joints. 
However, the improved performance of these VPTs comes with a heavy computation burden. 

\noindent \textbf{Efficient 3D HPE.}
Efficient 3D HPE is critical in computing resource-constrained environments. 
Existing explorations mainly focus on efficient architecture design \cite{simplebaseline,videopose,choi2021mobilehumanpose} and data redundancy reduction \cite{stride,pstmo,deciwatch,einfalt2023uplift}. 
VPose \cite{videopose} presents a fully convolutional architecture that processes multiple frames in parallel. 
Strided \cite{stride} designs a strided transformer encoder to aggregate redundant sequences. 
Recently, several studies \cite{pstmo,deciwatch,einfalt2023uplift} have attempted to improve model efficiency by uniformly sampling video sequences. 
For example, DeciWatch \cite{deciwatch} proposes a flow that takes sparsely sampled frames as inputs. 
However, this is suboptimal as it simply selects frames at a fixed interval in a static manner without considering their contextual cues. 
In contrast, we propose to utilize the cluster to dynamically select pose tokens of representative frames with high-level semantic representations. 
Besides, many efficient methods \cite{stride,einfalt2023uplift,poseformerv2} are designed for a specific model and none of them unifies the efficient design for different VPTs. 
We are the first to propose a plug-and-play framework for efficient VPTs, which can be plugged into common VPT models. 

\noindent \textbf{Token Pruning for Transformers.}
The self-attention complexity in transformers grows quadratically with the number of tokens, making it infeasible for high spatial or temporal resolution inputs. 
Many works 
\cite{dou2023tore,xie2022clustr,long2023beyond,kong2022spvit,chang2023making} attempt to alleviate this issue by using token pruning, which aims to select significant tokens from different inputs. 
They find that discarding less informative tokens in the deep transformer blocks only leads to a slight performance drop. 
DynamicViT \cite{rao2021dynamicvit} proposes a learnable prediction module to estimate the scores of tokens and prune redundant tokens. 
PPT \cite{ma2022ppt} selects important tokens based on the attention score. 
TCFormer \cite{zeng2022not} presents a token clustering transformer to cluster and merge tokens. 
In this work, we are the first to perform token pruning in VPTs for model acceleration. 
Unlike these studies that aim to reduce less related information (\textit{e.g.}, image background) from images in the spatial domain, we focus on reducing temporal redundancy by selecting a few pose tokens of representative frames in the temporal domain. 
Furthermore, we propose to restore the full-length temporal resolution to meet the domain-specific requirement of efficient video-based 3D HPE. 

\section{Method}
Figure~\ref{fig:overview} illustrates the overview of our Hourglass Tokenizer (HoT). 
Our HoT is a general-purpose \textit{pruning-and-recovering} framework that can use different token pruning and token recovering strategies (see Sec~\ref{sec:ablation}). 
For better token pruning and recovering, we propose token pruning cluster (TPC) and token recovering attention (TRA) modules and insert them into SOTA VPTs \cite{mhformer,mixste,motionbert}. 
Specifically, TPC takes the full-length pose tokens $x_{n} \in \mathbb{R}^{F \times J \times C}$ of $n$-th transformer block as inputs and outputs a few representative tokens $\tilde{x} \in \mathbb{R}^{f \times J \times C}$ ($f \ll F$), where $J$, $F$, and $f$ are the number of body joints, input frames, and representative tokens, respectively. 
Here, $C$ denotes the feature dimension. 
TRA recovers the full-length tokens from the tokens of the last transformer block $x_{L} \in \mathbb{R}^{f \times J \times C}$, where $L$ is the number of transformer blocks, resulting in recovered tokens $\hat{x} \in \mathbb{R}^{F \times J \times C}$. 
In the following section, we give details about the proposed TPC and TRA modules and show how to apply them to existing VPTs. 

\subsection{Token Pruning Cluster}
We observe that the existing VPTs \cite{{mhformer,mixste,motionbert}} take long video sequences as input and maintain the full-length sequence across all blocks (Figure~\ref{fig:pipeline} (a)), which is computationally expensive for high temporal resolution inputs. 
To tackle this issue, we propose to prune the pose tokens of video frames to improve the efficiency of VPTs. 
However, it is challenging to select a few pose tokens that maintain rich information for accurate 3D HPE. 

\begin{figure}[tb]
\centering
\includegraphics[width=1.00\linewidth]{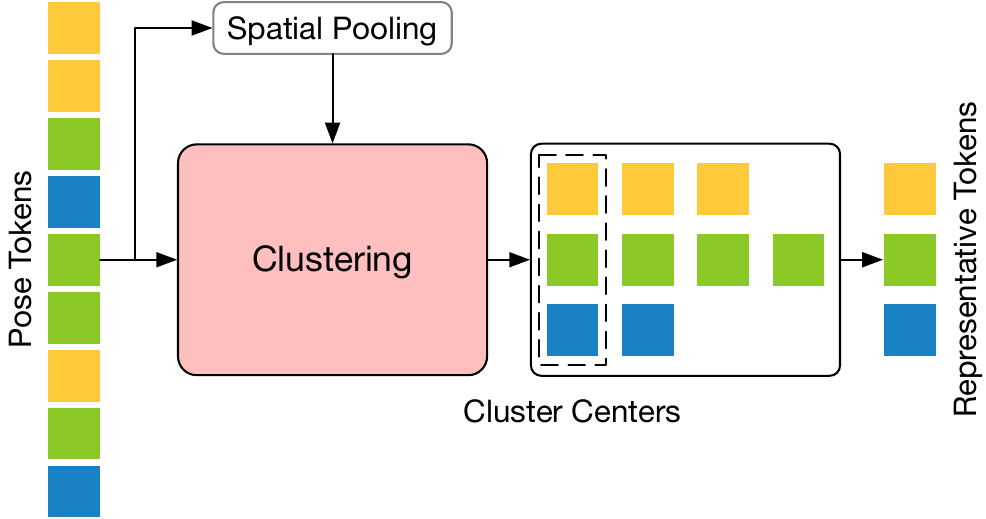}
\VspaceSF
\caption
{
Illustration of our token pruning cluster (TPC) architecture. 
Given the input pose tokens, we pool them in the spatial dimension, cluster the input tokens into several groups
according to the feature similarity of the resulting pooled tokens, and select the cluster centers as the representative tokens. 
}
\label{fig:tpc}
\VspaceL
\end{figure}

To address this challenge, we propose a simple, effective, and parameter-free token pruning cluster (TPC) that dynamically selects a few pose tokens of representative frames to eliminate video redundancy. 
The architecture of TPC is illustrated in Figure~\ref{fig:tpc}. 
Given the input pose tokens of $n$-th transformer blocks $x_{n} \in \mathbb{R}^{F \times J \times C}$, an average spatial pooling is used along the spatial dimension to remove spatial redundancy, resulting in pooled tokens $\overline{x}_{n} \in \mathbb{R}^{F \times C}$. 
Then, we apply an efficient density peaks clustering based on $k$-nearest neighbors (DPC-$k$NN) algorithm \cite{du2016study}. 
This algorithm clusters the input pose tokens into several groups according to the feature similarity of the pooled tokens without requiring an iterative process. 

The cluster centers of tokens are characterized by a higher density compared to their neighbors, as well as a relatively large distance from other tokens with higher densities. 
For a token $x^{i} \in \overline{x}_{n}$, the local density of tokens $\rho$ is calculated by:
\begin{equation}
    \rho_{i}=\exp (-\frac{1}{k} \sum_{x^{j} \in k\mathrm{NN}({x}^{i})} 
    \left\|x^i-x^j\right\|_{2}^{2}),
\end{equation}
where $k\mathrm{NN}\left({x}^{i}\right)$ are the $k$-nearest neighbors of a token $x^{i}$. 

We then define the $\delta_{i}$ that measures the minimal distance between the token $x^{i}$ and other tokens with higher density. 
The $\delta_{i}$ of the token with the highest density is set to the maximum distance between it and any other tokens. 
The $\delta_{i}$ of each token is calculated by:
\begin{equation}
    \delta_{i}=\left\{
    \begin{array}{l}
    \min _{j: \rho_j>\rho_i} \left\|x^i-x^j\right\|_{2}, \text { if } \exists \rho_{j}>\rho_{i} \\
    \max _{j} \left\|x^i-x^j\right\|_{2}, \text { otherwise }
    \end{array}.\right.
\end{equation}

The clustering center score of a token $x^{i}$ is denoted by combining the local density $\rho_i$ and minimal distance $\delta_i$ as $\rho_i \times \delta_i$. 
A higher score indicates that the token has both a large density and distance, showing a higher potential to be the cluster center.
The top-$f$-scored input pose tokens are selected as cluster centers, and the remaining tokens are assigned to the nearest cluster center with higher density. 

The cluster centers have high semantic diversity, containing more informative information than the other tokens. 
Therefore, the cluster centers serve as the representative tokens $\tilde{x} \in \mathbb{R}^{f \times J \times C}$ for efficient estimation, and the remaining tokens are discarded for reduction of video redundancy. 
Note that our method only prunes the tokens along the temporal dimension since the frame number $F$ is much larger than the joint number $J$ (\textit{e.g.}, $F{=}243$ and $J{=}17$), \textit{i.e.}, the expensive and redundant computational costs are dominated by the frame number in the temporal domain. 

\begin{figure}[tb]
\centering
\includegraphics[width=1.00\linewidth]{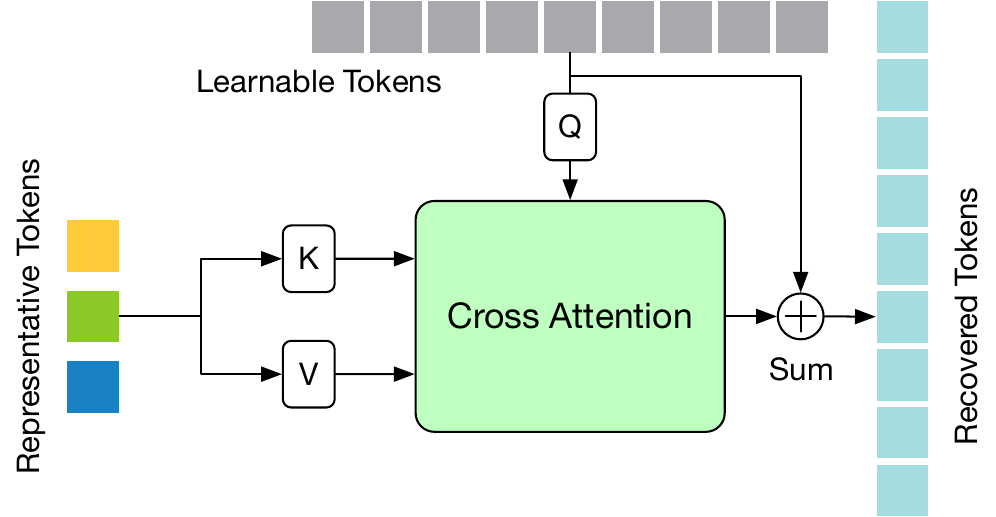}
\VspaceSF
\caption
{
Illustration of our token recovering attention (TRA) architecture. 
TRA takes the representative tokens of the last transformer block, along with learnable tokens that are initialized to zero, as input to recover the full-length tokens. 
}
\label{fig:tra}
\VspaceL
\end{figure}

\subsection{Token Recovering Attention}
A large number of pose tokens have been pruned by TPC, which significantly reduces the computational costs. 
However, for fast inference, a real-world 3D HPE system should be capable of estimating the consecutive 3D poses of all frames in a given video (this is called \textit{seq2seq} pipeline in \cite{mixste}). 
Therefore, different from some token pruning methods in vision transformers that can use a few selected tokens to directly perform classification \cite{rao2021dynamicvit,yin2022vit,marin2021token,liang2022not}, we need to recover the full-length tokens to keep the same number of tokens as the input video frames (in existing VPTs, each token corresponds to a frame). 
Meanwhile, for efficiency purposes, the recovering module should be lightweight. 

To this end, a lightweight token recovering attention (TRA) module is proposed to restore the spatio-temporal information from the selected pose tokens, as shown in Figure~\ref{fig:tra}. 
It only contains one multi-head cross-attention (MCA) layer without any additional networks. 
Formally, the dot-product attention \cite{transformer} in the MCA is defined as:
\begin{equation}
    \operatorname{Attention}(Q, K, V)=\operatorname{Softmax}\left(Q K^T / \sqrt{d}\right) V,
\end{equation}
where queries $Q \in \mathbb{R}^{n_{q} \times d}$, keys $K \in \mathbb{R}^{n_{k} \times d}$, and values $V \in \mathbb{R}^{n_{v} \times d}$. $d$ is the dimension and $\{n_{q}, n_{k},n_{v}\}$ are the number of tokens for $\{Q, K, V\}$, respectively. 

Our MCA takes the learnable tokens $x^{\prime} \in \mathbb{R}^{F \times C}$ that are initialized to zero as queries and the $j$-th joint representative tokens of the last transformer block $x_{L}^{j} \in \mathbb{R}^{f \times C}$ as keys and values, followed by a residual connection:
\begin{equation}
    \hat{x}^{j} = 
    x^{\prime} + 
    \operatorname{MCA}(x^{\prime}, x_{L}^{j}, x_{L}^{j}),
\end{equation}
where $\operatorname{MCA}(\cdot)$ is the function of MCA, and its inputs are queries, keys, and values. 
$\hat{x}^{j} \in \mathbb{R}^{F \times C}$ is the $j$-th joint recovered token, whose temporal dimension is the same as the queries (\textit{i.e.}, the designed learnable tokens). 

The TRA performs a reverse operation of selecting representative tokens, which recovers tokens of full-length temporal resolution from low ones using high-level spatio-temporal semantic information. 

\subsection{Applying to VPTs}
\label{sec:VPTs}
Recent studies of VPTs can be divided into two types of pipelines based on their inference outputs: \textit{seq2frame} \cite{poseformer,mhformer,stride,pstmo} and \textit{seq2seq} \cite{mixste,motionbert} pipelines. 
The \textit{seq2frame} pipeline outputs the 3D pose of the center frame and requires repeated inputs of 2D pose sequences with significant overlap to predict the 3D poses of all frames. 
This pipeline can achieve better performance by considering both past and future information, but it is not efficient due to repeated calculations. 
In contrast, the \textit{seq2seq} pipeline outputs 3D poses of all frames from the input 2D pose sequence at once, making it more efficient but leading to a degradation in performance. 
As a result, these two pipelines have their unique strengths, and we need to develop two strategies to better accommodate their different inference manners. 

For the \textit{seq2seq} pipeline, the outputs are all frames of the input video, and hence we need to restore the original temporal resolution. 
TPC and TRA are inserted into VPTs, where TPC prunes the tokens after a few transformer blocks and TRA recovers the full-length tokens after the last transformer block, as shown in Figure~\ref{fig:overview}. 
Specifically, given the input 2D pose sequence $p \in \mathbb{R}^{F \times J \times 2}$ detected by an off-the-shelf 2D HPE detector from a video, we first feed them into a pose embedding module to embed spatial and temporal information of pose frames, resulting in tokens $x \in \mathbb{R}^{F \times J \times C}$. 
The embedded tokens are then fed into a few transformer blocks. 
Next, the TPC selects a few representative tokens $\tilde{x} \in \mathbb{R}^{f \times J \times C}$, which are the inputs of subsequent transformer blocks. 
After the last transformer block, the TRA restores the original temporal resolution and produces recovered tokens $\hat{x} \in \mathbb{R}^{F \times J \times C}$. 
Finally, a regression head is added to estimate the 3D pose sequence $q \in \mathbb{R}^{F \times J \times 3}$. 

\begin{figure}[t]
\centering
\includegraphics[width=1.00\linewidth]{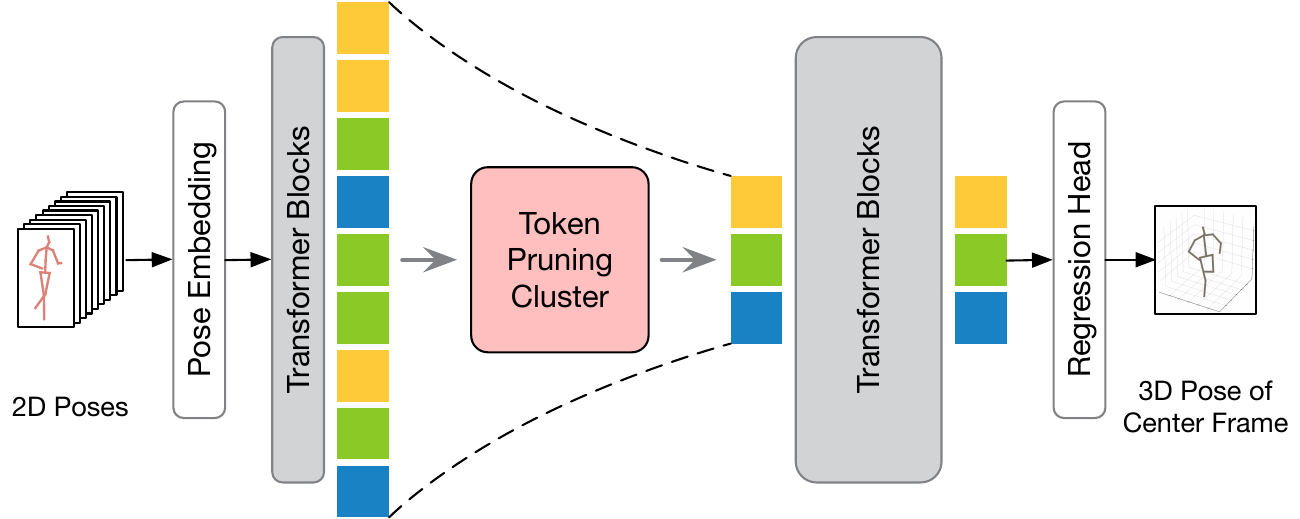}
\VspaceSF
\caption
{
Illustration of our framework on \textit{seq2frame} pipeline. 
The pose tokens are fed into TPC to select representative tokens. 
After the regression head, the 3D pose of the center frame is selected as the output for evaluation. 
}
\label{fig:seq2frame}
\end{figure}

For the \textit{seq2frame} pipeline, the output is the 3D pose of the center frame. 
Therefore, TRA is unnecessary and we only insert TPC into VPTs. 
Since the token of the center frame directly corresponds to the output and can provide crucial information to the final estimation, we concatenate it with the selected tokens to make this pipeline work better. 
As shown in Figure~\ref{fig:seq2frame}, the early stages of both pipelines share the same workflow. 
After the last transformer block, the tokens are directly sent to the regression head to perform regression and the 3D pose of center frame $q_{center} \in \mathbb{R}^{1 \times J \times 3}$ is selected as the final prediction. 

\section{Experiments}
\subsection{Datasets and Evaluation Metrics}
\noindent \textbf{Datasets.}
We evaluate our method on two 3D HPE benchmark datasets: Human3.6M \cite{ionescu2013human3} and MPI-INF-3DHP \cite{mehta2017monocular}. 
Human3.6M is the most widely used dataset for 3D HPE. 
It consists of 3.6 million video frames recorded by four RGB cameras at 50 Hz in an indoor environment. 
This dataset includes 11 actors performing 15 daily actions. 
Following \cite{zhao2019semantic,gong2021poseaug,zou2021modulated,li2023multi}, subjects S1, S5, S6, S7, S8 are used for training and subjects S9, S11 are used for testing. 
MPI-INF-3DHP is another popular 3D HPE dataset. 
This dataset contains 1.3 million frames collected in indoor and outdoor scenes. 
It is smaller than Human3.6M but more challenging due to its diverse scenes, viewpoints, and motions. 

\noindent \textbf{Evaluation Metrics.}
For Human3.6M, we use the most commonly used mean per joint position error (MPJPE) as the evaluation metric, which measures the average Euclidean distance between estimated and ground truth 3D joint coordinates in millimeters. 
For MPI-INF-3DHP, we follow previous works \cite{poseformer,mhformer,mixste} to report metrics of MPJPE, percentage of correct keypoint (PCK) with the threshold of 150mm, and area under curve (AUC). 

\subsection{Implementation Details}
\label{sec:Implementation Details}
The network is implemented using the PyTorch framework on one consumer-level NVIDIA RTX 3090 GPU with 24G memory. 
Our method builds upon MHFormer \cite{mhformer}, MixSTE \cite{mixste}, and MotionBERT \cite{motionbert} for their largest frame number (\textit{i.e.}, $F{=}351,243,243$) models. 
For a speed-accuracy trade-off, by default, we set \{$F{=}351$, $n{=}1$, $f{=}117$\} for MHFormer, \{$F{=}243$, $n{=}3$, $f{=}81$\} for MixSTE, and \{$F{=}243$, $n{=}1$, $f{=}81$\} for MotionBERT. 
Note that MHFormer is designed for \textit{seq2frame} pipeline, so we only implement our TPC on it. 
MixSTE and MotionBERT are designed for \textit{seq2seq} pipeline and can be implemented on both \textit{seq2frame} (with TPC) and \textit{seq2seq} (with HoT) pipelines. 

\begin{table}[t]
\scriptsize
\centering
\caption
{ 
Comparison of efficiency and accuracy between \textit{seq2seq} ($*$) and \textit{seq2frame} ($\dagger$) inference pipelines. 
Frame per second (FPS) was computed on a single GeForce RTX 3090 GPU. 
}
\VspaceST
\setlength{\tabcolsep}{0.30mm}
\begin{tabular}{l|cll|c}
\toprule [1pt]
Method &Param (M) &FLOPs (G) &FPS &MPJPE $\downarrow$  \\

\midrule [0.5pt]

MixSTE \cite{mixste} ($*$) &33.78 &277.25 &10432 &40.9 \\

\blue{HoT} w. MixSTE ($*$) &35.00 &167.52 (\textbf{\blue{$\downarrow$ 39.6\%}}) &15770 (\textbf{\blue{$\uparrow$ 51.2\%}})  &{41.0} \\

\midrule [0.5pt]

MixSTE \cite{mixste} ($\dagger$)  &33.78 &277.25 &43 &40.7 \\

\blue{TPC} w. MixSTE ($\dagger$) &33.78 &161.73 (\textbf{\blue{$\downarrow$ 41.7\%}}) &68 (\textbf{\blue{$\uparrow$ 58.1\%}}) &{40.4} \\

\midrule [0.5pt]
\midrule [0.5pt]
MotionBERT \cite{motionbert} ($*$) &16.00 &131.09 &14638 &39.8 \\

\blue{HoT} w. MotionBERT ($*$) &16.35 &63.21 (\textbf{\blue{$\downarrow$ 51.8\%}}) &25526 (\textbf{\blue{$\uparrow$ 74.4\%}}) &{39.8} \\

\midrule [0.5pt]

MotionBERT \cite{motionbert} ($\dagger$) &16.00 &131.09 &60 &39.5 \\

\blue{TPC} w. MotionBERT 
($\dagger$)&16.00 &61.04 (\textbf{\blue{$\downarrow$ 53.4\%}}) &109 (\textbf{\blue{$\uparrow$ 81.7\%}}) &{39.2} \\

\bottomrule [1pt]

\end{tabular}
\label{table:pipeline}
\VspaceL
\end{table}

\subsection{Ablation Study}
\label{sec:ablation}
To validate the effectiveness of our method, we conduct extensive ablation studies on Human3.6M dataset. 

\noindent \textbf{Inference Pipeline.}
In Table~\ref{table:pipeline}, we compare the efficiency and accuracy between different inference pipelines (mentioned in Sec~\ref{sec:VPTs}). 
We conduct experiments on MixSTE \cite{mixste} and MotionBERT \cite{motionbert} because both are designed for \textit{seq2seq} pipeline and can be evaluated on both \textit{seq2frame} and \textit{seq2seq} pipelines. 
As shown in the table, the \textit{seq2frame} can achieve better estimation accuracy by taking advantage of past and future information but lower efficiency due to repeated computations, \textit{e.g.}, 40.7mm vs. 40.9mm and 43 FPS vs. 10432 FPS for MixSTE (about 243$\times$ lower). 
As our TPC is parameter-free and TRA is lightweight, our method with TPC introduces no additional parameters, and HoT w. MotionBERT only introduces additional 0.35M (2.2\%) parameters, which can be neglected. 
Moreover, our method reduces the computational costs and improves the inference speed of these two pipelines, while maintaining or obtaining better performance. 

For the \textit{seq2seq}, our method can reduce the FLOPs of MixSTE and MotionBERT by 39.6\% and 51.8\% and improve the FPS by 51.2\% and 74.4\%, while estimation errors only drop 0.1mm (0.24\%) and remain unchanged, respectively. 
For the \textit{seq2frame}, our TPC w. MixSTE can reduce the FLOPs by 41.7\% and improve the FPS by 58.1\%, while bringing 0.3mm improvement. 
Additionally, our TPC w. MotionBERT can reduce 53.4\% FLOPs and improve 81.7\% FPS, while the estimation errors are reduced from 39.5mm to 39.2mm. 
Note that our method with TPC outperforms the one utilizing HoT. 
This is reasonable since our TRA in HoT is a reverse operation that uses inadequate information to recover the full-length tokens. 
In the following ablations, we take these two inference pipelines into account to sufficiently explore the proposed method, and we choose MixSTE \cite{mixste} as the baseline since it is the first \textit{seq2seq} transformer-based architecture and MotionBERT \cite{motionbert} is its follow-ups. 

\begin{table}[t]
\footnotesize
\centering
\caption
{ 
Ablation study on the block index of representative tokens ($n$) under the \textit{seq2frame} pipeline. 
Here, $^{*}$ denotes the result without re-training.
}  
\VspaceST
\setlength{\tabcolsep}{0.40mm}
\begin{tabular}{l|cl|cc}
\toprule [1pt]
Method &Param (M) &FLOPs (G) &MPJPE$^{*}$ &MPJPE $\downarrow$  \\
\midrule [0.5pt]

MixSTE \cite{mixste} &33.78 &277.25 &40.7 &40.7 \\
\midrule [0.5pt]

\blue{TPC} w. MixSTE, $n{=}2$ &33.78 &121.52 (\textbf{\blue{$\downarrow$ 56.2\%}}) &41.2 &40.7 \\

\blue{TPC} w. MixSTE, $n{=}3$ &33.78 &147.47 (\textbf{\blue{$\downarrow$ 46.8\%}}) &41.2 &40.5 \\

\blue{TPC} w. MixSTE, $n{=}5$ &33.78 &199.38 (\textbf{\blue{$\downarrow$ 28.1\%}}) &40.9 &40.2 \\

\blue{TPC} w. MixSTE, $n{=}7$ &33.78 &251.29 (\textbf{\blue{$\downarrow$ 09.4\%}}) &\textbf{\red{40.7}} &\textbf{\red{39.9}} \\

\bottomrule [1pt]
\end{tabular}
\label{table:block}
\VspaceS
\end{table}

\begin{table}[t]
\footnotesize
\centering
\caption
{ 
    Ablation study on the number of representative tokens ($f$) under the \textit{seq2seq} pipeline. 
}
\VspaceST
\setlength{\tabcolsep}{1.40mm}
\begin{tabular}{l|cl|c}
\toprule [1pt]
Method &Param (M) &FLOPs (G) &MPJPE $\downarrow$  \\
\midrule [0.5pt]

MixSTE \cite{mixste} &33.78 &277.25 &40.9 \\

\midrule [0.5pt]

\blue{HoT} w. MixSTE, $f {=} 9$ &34.96 &114.90 (\textbf{\blue{$\downarrow$ 58.6\%}}) &{43.5} \\

\blue{HoT} w. MixSTE, $f {=} 16$ &34.97 &120.01 (\textbf{\blue{$\downarrow$ 56.7\%}}) &{42.2} \\

\blue{HoT} w. MixSTE, $f {=} 61$ &34.99 &152.90 (\textbf{\blue{$\downarrow$ 44.9\%}}) &{41.2} \\ 

\blue{HoT} w. MixSTE, $f {=} 81$ &35.00 &167.52 (\textbf{\blue{$\downarrow$ 39.6\%}}) &\textbf{\red{41.0}} \\

\blue{HoT} w. MixSTE, $f {=} 135$ &35.03 &206.98 (\textbf{\blue{$\downarrow$ 25.3\%}}) &{41.3} \\

\bottomrule [1pt]
\end{tabular}
\label{table:number}
\VspaceL
\end{table}

\begin{figure*}[tb]
\centering
\includegraphics[width=1.00\linewidth]{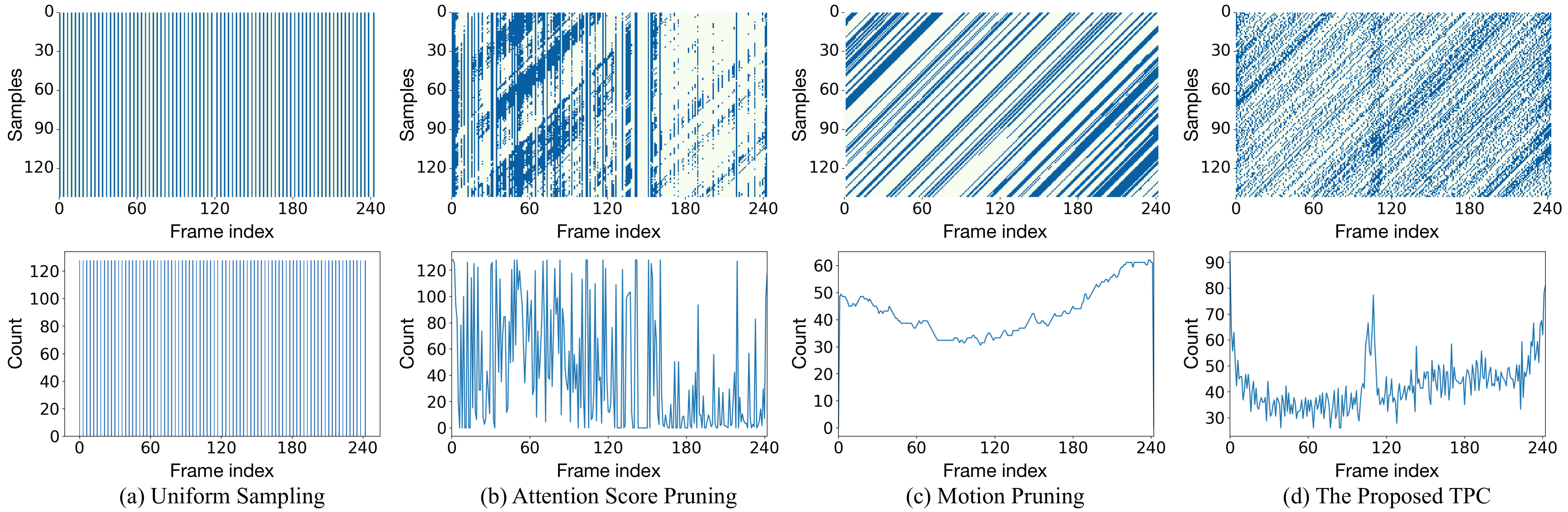}
\VspaceSF
\caption
{
Statistics visualization of selected tokens for different token pruning strategies. 
\textbf{Top}: Frame indexes of selected tokens for some samples (140 samples) of consecutive video sequences (243 frames). 
Blue points are selected tokens and white points are pruned tokens. 
\textbf{Bottom}: Frequency count of frame indexes of selected tokens for these samples. 
}
\label{fig:index}
\VspaceL
\end{figure*}

\noindent \textbf{Block index of Representative Tokens.}
The TPC can be inserted into optional transformer blocks, thereby adjusting the trade-off between computational costs and performance on demand in a flexible manner. 
Table~\ref{table:block} studies this under \textit{seq2frame} pipeline ($f$ is fixed to 61). 
Since TPC is a data-dependent scheme that introduces no extra parameters and transformers are input agnostic \cite{bolya2022token,fayyaz2022adaptive}, we can evaluate models with or without re-training. 
Increasing the block index of representative tokens can reduce the estimation error, but it also leads to higher computational costs. 
This indicates the deeper blocks of transformers contain more redundancy while the shallower blocks retain more useful information.  
Our method achieves competitive results without re-training while reducing FLOPs. 
When it works with re-training (training from scratch without pre-trained models), our method attains better performance. 
Our TPC w. MixSTE ($n {=} 2$) achieves the same results while reducing 56.2\% FLOPs and TPC w. MixSTE ($n {=} 7$) improves the performance from 40.7mm to 39.9mm while reducing 9.4\% FLOPs. 

\noindent \textbf{Number of Representative Tokens.}
The number of representative tokens $f$ can also be flexibly adjusted. 
In Table~\ref{table:number}, we fix $n$ to 3 and vary $f$ under \textit{seq2seq} pipeline. 
Increasing $f$ can reduce the FLOPs, but the best performance is achieved by using $f{=}81$. 
The reason for this is that an appropriate number of representative tokens can bring a good trade-off between retaining important information and reducing redundant information for both the pruning and recovering stages. 
Therefore, the optimal hyper-parameters for our HoT w. MixSTE are $n{=}3$ and $f{=}81$. 

\begin{table}[t]
\Huge
\centering
\caption
{ 
Ablation study on the design choices of token pruning. 
``FN'' denotes the frame noise that calculates the MPJPE of selected frames. 
``Full'', ``Pruned'', ``Selected'', and ``Center'' denote the MPJPE of all frames, pruned frames, selected frames, and the center frame, respectively. 
}  
\VspaceST
\resizebox{\columnwidth}{!}{
\begin{tabular}{lcccccc}
\toprule [3pt]
&  & \multicolumn{3}{c}{\textit{seq2seq}} & \multicolumn{2}{c}{\textit{seq2frame}} \\
\cmidrule(lr){2-5}\cmidrule(lr){6-7}
Method &FN &Full $\downarrow$  &Pruned $\downarrow$  &Selected $\downarrow$  &Center $\downarrow$  &Selected $\downarrow$   \\
\midrule [2.0pt]
MixSTE \cite{mixste} &6.61 &40.9 &- &- &40.7 &- \\ 
\midrule [2.0pt]

Ours, Uniform Sampling &6.61 &41.4 &41.3 &41.4 &40.7 &40.8 \\ 

Ours, Attention Pruning &6.56 &42.1 &42.5 &41.5 &42.3 &44.4  \\ 

Ours, Motion Pruning &7.00 &42.8 &43.4 &41.6 &41.3 &42.3 \\ 

Ours, the Proposed TPC &6.63 &\textbf{\red{41.0}} &\textbf{\red{41.3}} &\textbf{\red{40.2}} &\textbf{\red{40.4}} &\textbf{\red{39.4}} \\ 

\bottomrule [3pt]
\end{tabular}
}
\label{table:pruning}
\VspaceL
\vspace{-0.30cm}
\end{table}

\noindent \textbf{Token Pruning Design.}
Our HoT is a general-purpose \textit{pruning-and-recovering} framework that can be equipped with different token pruning and recovering strategies. 
In Table~\ref{table:pruning}, we compare different token pruning strategies, including attention pruning \cite{wang2022kvt,ma2022ppt}, uniform sampling \cite{pstmo,deciwatch,einfalt2023uplift}, and motion pruning that selects tokens with top-$k$-large motions. 
To measure the quality of selected tokens, we define a frame noise metric, which calculates the MPJPE of the 2D poses of input frames corresponding to the selected indexes. 
As the table shows, the frame noise values among these methods are similar (around 6.6mm) except for the motion pruning (7.0mm). 
This is because selecting tokens with top-$k$-large motion introduces some noise frames that differ significantly from clean frames, which can adversely affect performance. 
Moreover, our proposed TPC
outperforms all other token pruning strategies, particularly for selected frames. 
Our TPC outperforms the uniform sampling strategy by 1.2mm (40.2mm vs. 41.4mm) and 1.4mm (39.4mm vs. 40.8mm) for selected frames under \textit{seq2seq} and \textit{seq2frame} pipelines, respectively. 
This emphasizes that the 3D pose results of our selected frames are easier to estimate, and our method can select more representative frames from a video. 

\begin{table}[t]
\Huge
\centering
\caption
{
Ablation study on the design choices of token recovering. 
$\Delta$ represents the performance gap between the results of pruned frames and selected frames.
}
\VspaceST
\resizebox{\columnwidth}{!}{

\begin{tabular}{l|cc|ccc|c}
\toprule [3pt]
Method &Param &FLOPs &Full $\downarrow$  &Pruned $\downarrow$  &Selected $\downarrow$  &$\Delta$ \\
\midrule [2.0pt]
MixSTE \cite{mixste} &33.78 &277.25 &40.9 &- &- &- \\ 
\midrule [2.0pt]

Ours, Nearest Interpolation &33.78 &161.73 &41.5 &42.2 &40.2 &2.0  \\ 

Ours, Linear Interpolation &33.78 &161.73 &41.3 &41.9 &\textbf{\red{40.0}} &1.9  \\ 

Ours, the Proposed TRA &35.00 &167.52 &\textbf{\red{41.0}} &\textbf{\red{41.3}} &40.2 &\textbf{\red{1.1}}  \\

\bottomrule [3pt]
\end{tabular}
}
\label{table:recover}
\VspaceL
\vspace{-0.30cm}
\end{table}

Furthermore, we statistically visualize selected tokens of these four token pruning strategies. 
For better observation, we take samples of consecutive video sequences as input with a temporal interval of 1 between neighboring samples. 
The frame indexes and the frequency count of frame indexes of the selected tokens are shown in Figure~\ref{fig:index} (top) and Figure~\ref{fig:index} (bottom). 
Uniform sampling and motion pruning are static pruning methods because the former selects tokens at a fixed frame interval (equidistance in the top of Figure~\ref{fig:index} (a)), while the latter selects tokens with the top-$k$-large motions that move with the input sequence (oblique triangle in the top of Figure~\ref{fig:index} (c)). 
Instead, the attention score pruning and our method are dynamic methods that consider the significance of input tokens. 
The bottom of Figure~\ref{fig:index}~(b) shows that attention score pruning tends to select tokens in the left half of a sequence, indicating that the selected tokens tend to be similar to each other \cite{wang2022vtc} and thus lack diversity.
Our method primarily selects tokens at the beginning, center, and end of a sequence (the bottom of Figure~\ref{fig:index}~(d)). 
This is reasonable since these three parts can represent the rough motion of an entire sequence, which contributes a lot to accurate estimation. 
These findings highlight that our method not only eliminates the redundancy of video frames but also selects tokens with high semantic diversity (the top of Figure~\ref{fig:index} (d) appears to be irregular), thus selecting more representative pose tokens for more accurate estimation. 

\begin{table}[t]
\footnotesize
\centering
\caption
{ 
Comparison of parameters (M), FLOPs (G), and MPJPE with SOTA VPTs on Human3.6M. 
Here, $F$ denotes the number of input frames. 
$^{*}$ indicates our re-implementation.
}  
\VspaceST
\setlength{\tabcolsep}{0.60mm} 
\begin{tabular}{l|cclc}
\toprule [1pt]
Method &$F$ &Param &FLOPs &MPJPE $\downarrow$  \\
\midrule [0.5pt]
PoseFormer (ICCV'21) \cite{poseformer} &81 &9.60 &1.63 &44.3 \\

Strided (TMM'22) \cite{stride} &351 &4.35 &1.60 &43.7 \\

P-STMO (ECCV'22) \cite{pstmo} &243 &7.01 &1.74 &42.8 \\

STCFormer (CVPR'23) \cite{tang20233d} &243 &18.93 &156.22 &40.5 \\

\midrule [0.5pt]

MHFormer (CVPR'22) \cite{mhformer} &351 &31.52 &14.15 &43.0 \\

\blue{TPC} w. MHFormer (Ours) &351 &31.52 &8.22 (\textbf{\blue{$\downarrow$ 41.91\%}}) &{43.0} \\

\midrule [0.5pt]

MixSTE (CVPR'22) \cite{mixste} &243 &33.78 &277.25 &40.9 \\

\blue{HoT} w. MixSTE (Ours) &243 &35.00 &167.52 (\textbf{\blue{$\downarrow$ 39.6\%}}) &{41.0} \\

\blue{TPC} w. MixSTE (Ours) &243 &33.78 &251.29 (\textbf{\blue{$\downarrow$ 09.4\%}}) &39.9 \\

\midrule [0.5pt]

MotionBERT (ICCV'23) \cite{motionbert} &243 &16.00 &131.09 &39.2 \\

MotionBERT (ICCV'23) \cite{motionbert}$^{*}$ &243 &16.00 &131.09 &39.8 \\

\blue{HoT} w. MotionBERT (Ours) & 243 &16.35 &63.21 (\textbf{\blue{$\downarrow$ 51.8\%}}) &39.8 \\

\blue{TPC} w. MotionBERT (Ours) & 243 &16.00 &91.38 (\textbf{\blue{$\downarrow$ 30.3\%}}) &\textbf{\red{39.0}} \\

\bottomrule [1pt]
\end{tabular}
\label{table:h36m}
\VspaceL
\end{table}

\noindent \textbf{Token Recovering Design.}
The token recovering strategies in our HoT can also be designed in different manners, as studied in Table~\ref{table:recover}. 
It shows that linear and nearest interpolation operations are parameter-free and achieve competitive results due to data redundancy (\textit{i.e.}, nearby poses are similar) on Human3.6M (captured by 50 Hz cameras). 
Our TRA achieves better performance while introducing negligible parameters and FLOPs. 
These results validate the effectiveness of the proposed TRA, 
highlighting the benefits of using high-level semantic information for pose token recovering. 
Besides, the experiments show that the proposed TRA achieves the lowest performance gap between the estimated 3D poses of pruned frames and selected frames. 
This further demonstrates the effectiveness of our TRA, which can recover more accurate results based on the limited information provided by the selected tokens. 

\subsection{Comparison with state-of-the-art methods}
\noindent \textbf{Human3.6M.}
Current SOTA performance on Human3.6M is achieved by transformer-based architectures. 
We compare our method with them by adding it to three very recent VPTs: 
MHFormer \cite{mhformer}, MixSTE \cite{mixste}, and MotionBERT \cite{motionbert}. 
These three models significantly outperform previous works at the cost of high computational complexity, thus we choose them as baselines to evaluate our method. 
The comparisons are shown in Figure~\ref{fig:flops} and Table~\ref{table:h36m}. 
We report the results of TPC w. MixSTE with \{$n{=}7$, $f{=}61$\} and TPC w. MotionBERT with \{$n{=}2$, $f{=}121$\}. 
As shown in the table, our method can reduce the computational costs of recent VPTs while maintaining the ability of the model. 
For example, our HoT w. MotionBERT saves 51.8\% FLOPs while maintaining accuracy, and our TPC w. MotionBERT obtains better performance with 0.8mm improvements while reducing computational costs by 30.3\% in FLOPs. 
These results demonstrate the effectiveness and efficiency of our method, while also revealing that existing VPTs incur redundant computational costs that contribute little to the estimation accuracy or even decrease the accuracy. 
In addition, our method can remove these unnecessary computational costs while achieving comparable or even superior performance. 

\begin{table}[t]
\footnotesize
\centering
\caption
{
Quantitative comparison with SOTA methods on MPI-INF-3DHP. 
}
\VspaceST
\setlength{\tabcolsep}{2.20mm} 
\begin{tabular}{@{}l|ccc@{}}
\toprule [1pt]
Method & PCK $\uparrow$  & AUC $\uparrow$  & MPJPE $\downarrow$  \\
\midrule [0.5pt]
VPose (CVPR'19) \cite{videopose} ($F {=} 81$) &86.0 &51.9 &84.0 \\
UGCN (ECCV'20) \cite{wang2020motion} ($F {=} 96$) &86.9 &62.1 &68.1 \\
Anatomy3D (TCSVT'21) \cite{chen2021anatomy} ($F {=} 81$) &87.9 &54.0 &78.8 \\
PoseFormer (ICCV'21) \cite{poseformer} ($F {=} 9$) &{88.6} &{56.4} &{77.1} \\ 

\midrule [0.5pt]

MHFormer (CVPR'22) \cite{mhformer} ($F {=} 9$) &{93.8} &{63.3} &{58.0} \\

\blue{TPC} w. MHFormer (Ours, $F {=} 9$) &94.0 &63.3 &58.4 \\

\midrule [0.5pt]

MixSTE (CVPR'22) \cite{mixste} ($F {=} 27$) &{94.4} &{66.5} &{54.9} \\

\blue{HoT} w. MixSTE (Ours, $F {=} 27$) &\textbf{\red{94.8}} &\textbf{\red{66.5}} &\textbf{\red{53.2}} \\

\bottomrule [1pt]
\end{tabular}
\label{table:3dhp}
\VspaceL
\end{table}

\noindent \textbf{MPI-INF-3DHP.}
We further evaluate our method on MPI-INF-3DHP dataset in Table~\ref{table:3dhp}. 
For a fair comparison, following \cite{mhformer,mixste}, we implement our method on MHFormer with \{$F{=}9$, $n{=}1$, $f{=}3$\} and MixSTE with \{$F{=}27$, $n{=}3$, $f{=}9$\}. 
It can be found that our method (TPC w. MHFormer and HoT w. MixSTE) achieves competitive performance, demonstrating the effectiveness of our method in both indoor and outdoor scenes. 
Besides, our method can also work well with a small temporal receptive field. 

\section{Conclusion}
This paper presents Hourglass Tokenizer (HoT), a plug-and-play \textit{pruning-and-recovering} framework for efficient transformer-based 3D human pose estimation from videos. 
Our method reveals that maintaining the full pose sequence is unnecessary, and using a few pose tokens of representative frames can achieve both high efficiency and estimation accuracy. 
Comprehensive experiments demonstrate that our method is compatible and general. 
It can be easily incorporated into common VPT models on both \textit{seq2seq} and \textit{seq2frame} pipelines while effectively accommodating various token pruning and recovery strategies, thereby highlighting its potential for using future ones. 
We hope HoT can enable the creation of stronger and faster VPTs. 

{\small
\noindent \textbf{Acknowledgements.} This work was supported by the National Natural Science Foundation of China (No. 62073004), Natural Science Foundation of Shenzhen (No. JCYJ20230807120801002), and by the MUR PNRR project FAIR (PE00000013) funded by the NextGenerationEU. 
}

{\small
\bibliographystyle{ieeenat_fullname}
\bibliography{ref}
}

\clearpage
\appendix
{\noindent\Large\textbf{Supplementary Material}}
\newline

This supplementary material covers  the following details:
\begin{itemize}
\item A brief description of video pose transformers (Sec.~\ref{sec:Video Pose Transformers}). 
\item Computation complexity of transformers (Sec.~\ref{sec:Computation Complexity}). 
\item Additional implementation details (Sec.~\ref{sec:Implementation Details}). 
\item Additional quantitative results (Sec.~\ref{sec:Quantitative Results}). 
\item Additional ablation studies (Sec.~\ref{sec:Ablation Study})
\item Additional visualization results (Sec.~\ref{sec:Visualization Results}). 
\end{itemize}

\section{Video Pose Transformers}
\label{sec:Video Pose Transformers}
\begin{figure}[t]
\centering
\includegraphics[width=1.00\linewidth]{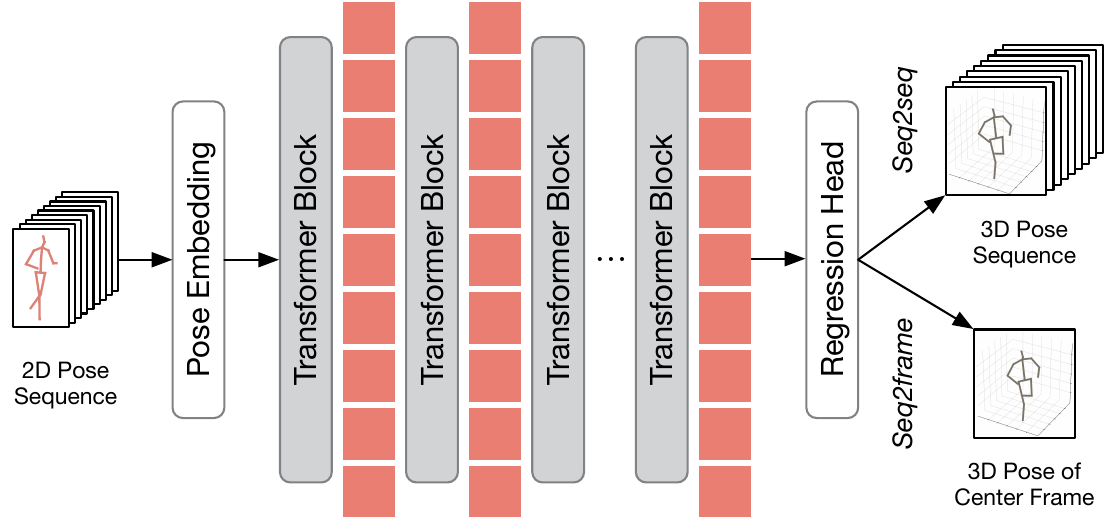}
\caption
{
Summary of VPT architectures. 
Existing VPTs typically contain a pose embedding module, a stack of transformer blocks, and a regression head module. 
The outputs of the regression head can be either the 3D poses of all frames for the \textit{seq2seq} pipeline or the 3D pose of the center frame for the \textit{seq2frame} pipeline.
}
\label{fig:vpt_supp}
\VspaceSUPPS
\end{figure}

Recent studies of video pose transformers (VPTs) \cite{poseformer,stride,li2023multi,mixste,pstmo,motionbert} are mainly designed to estimate 3D poses from 2D pose sequences. 
These VPTs share a similar architecture, which includes a pose embedding module (often containing only a linear layer) to embed spatial and temporal information of pose sequences, a stack of transformer blocks to learn global spatio-temporal correlations, and a regression module to predict 3D human poses. 
We summarize the architecture in Figure~\ref{fig:vpt_supp}. 
There are two types of pipelines based on their outputs: the \textit{seq2frame} pipeline outputs the 3D poses of all frames, while the \textit{seq2seq} pipeline outputs the 3D pose of the center frame. 

\section{Computation Complexity}
\label{sec:Computation Complexity}
Each transformer block consists of a multi-head self-attention (MSA) layer and a feed-forward network (FFN) layer. 
Let $N$ be the number of tokens, $D$ be the dimension, and $2D$ be the expanding dimension in the FFN (the expanding ratio in VPTs is typically 2). 
The calculational costs of MSA and FFN are $\mathcal{O}\left(4 N D^2+2 N^2 D\right)$, and $\mathcal{O}\left(4 N D^2\right)$, respectively. 
Thus, the total computational complexity is $\mathcal{O}\left(8 N D^2+2 N^2 D\right)$, which makes VPTs computationally expensive. 
Since the dimension $D$ is important to determine the modeling ability and most recent VPTs employ a $D$ of 512 or 256, we follow their hyperparameter settings and propose to prune pose tokens of video frames (\textit{i.e.}, reducing $N$) to reduce the computational cost of VPTs. 

\section{Additional Implementation Details}
\label{sec:Implementation Details}
Our method is built upon three very recent VPTs: MHFormer \cite{mhformer}, MixSTE \cite{mixste}, and MotionBERT \cite{motionbert}. 
These VPTs achieve state-of-the-art performance but are computationally expensive compared to previous methods (see Table~\ref{table:h36m}). 
We choose these VPTs as baselines to evaluate our method, which focuses on preserving the ability to model spatio-temporal dependencies while reducing computational costs. 
We adopt most of the optimal hyperparameters and training strategies used in \cite{mhformer,mixste,motionbert}, as shown in Table \ref{table:details_supp}. 
We also use the same loss functions for training, such as MPJPE loss for MHFormer, and weighted MPJPE loss, temporal consistency loss (TCLoss), and mean per-joint velocity error (MPJVE) for MixSTE. 

\begin{table}[t]
\footnotesize
\centering
\caption
{ 
    Implementation details of our method on MHFormer \cite{mhformer}, MixSTE \cite{mixste}, and MotionBERT \cite{motionbert}. 
    ($L$) - number of transformer blocks, 
    ($C$) - dimension, 
    (LR) - initial learning rate,
    (Flip) - horizontal flip augmentation,
    (CPN) - Cascaded Pyramid Network \cite{li2020cascaded},
    (SH) - Stack Hourglass \cite{newell2016stacked}. 
}
\setlength{\tabcolsep}{2.0mm}
\begin{tabular}{l|cccc|c}
\toprule [1pt]
Config &MHFormer \cite{mhformer} &MixSTE \cite{mixste} &MotionBERT \cite{motionbert} \\
\midrule [0.5pt]

$L$ &3 &8 &5 \\

$C$ &512 &512 &256 \\

Training Epoch &20 &160 &120 \\

Batch Size &210 &4 &4 \\

LR &$1 {\times} 10^{-3}$ &$4 {\times} 10^{-5}$ &$5 {\times} 10^{-4}$ \\

Optimizer &Amsgrad &Adam &Adam \\

Augmentation &Flip &Flip &Flip  \\

2D Detector &CPN &CPN &SH \\

\bottomrule [1pt]
\end{tabular}
\label{table:details_supp}
\VspaceSUPP
\end{table}

Since our TRA is designed for \textit{seq2seq} pipeline, it is unnecessary to add it to the model which is designed for \textit{seq2frame} pipeline (\textit{e.g.}, MHFormer). 
To provide a comprehensive analysis of our method, we report results with TPC and with both TPC and TRA. 
We denote the resulting models as follows:
\begin{itemize}
    \item \blue{HoT} w. MixSTE (MixSTE + TPC + TRA),
    \item \blue{HoT} w. MotionBERT (MotionBERT + TPC + TRA),
\end{itemize}
which are designed for \textit{seq2seq} pipeline, and: 
\begin{itemize}
    \item \blue{TPC} w. MHFormer (MHFormer + TPC),
    \item \blue{TPC} w. MixSTE (MixSTE + TPC),
    \item \blue{TPC} w. MotionBERT (MotionBERT + TPC),
\end{itemize}
which are designed for \textit{seq2frame} pipeline. 

\begin{table}[t]
\footnotesize
\centering
\caption
{ 
    Comparison of GPU memory cost (G) and training time (min/epoch) on a single  GeForce RTX 3090 GPU. 
} 
\setlength{\tabcolsep}{1.45mm}
\begin{tabular}{l|llc}
\toprule [1pt]
Method &GPU Memory &Training Time &MPJPE $\downarrow$ \\
\midrule [0.5pt]

MHFormer \cite{mhformer} &24.1 &223.2 &43.0 \\
\blue{TPC} w. MHFormer &13.8 (\textbf{\blue{$\downarrow$ 42.7\%}}) &131.0 (\textbf{\blue{$\downarrow$ 39.7\%}}) &43.0 \\

\midrule [0.5pt]

MixSTE \cite{mixste} &11.4 &17.0 &40.9 \\
\blue{HoT} w. MixSTE &7.6 (\textbf{\blue{$\downarrow$ 33.3\%}}) &10.5 (\textbf{\blue{$\downarrow$ 38.2\%}}) &41.0 \\
\blue{TPC} w. MixSTE &7.3 (\textbf{\blue{$\downarrow$ 36.0\%}}) &10.1 (\textbf{\blue{$\downarrow$ 40.6\%}}) &40.4 \\

\midrule [0.5pt]

MotionBERT \cite{motionbert} &10.7  &17.4 &39.8 \\
\blue{HoT} w. MotionBERT &6.1 (\textbf{\blue{$\downarrow$ 43.0\%}}) &8.9 (\textbf{\blue{$\downarrow$ 47.5\%}}) &39.8 \\
\blue{TPC} w. MotionBERT &5.7 (\textbf{\blue{$\downarrow$ 46.7\%}}) &8.4 (\textbf{\blue{$\downarrow$ 51.7\%}}) &\textbf{\red{39.2}}  \\

\bottomrule [1pt]
\end{tabular}
\label{table:memory_supp}
\VspaceSUPPS
\end{table}

\begin{table}[t]
\footnotesize
\centering
\caption
{ 
    Comparison with MixSTE. 
}
\setlength{\tabcolsep}{1.7mm}
\begin{tabular}{l|cccc|c}
\toprule [1pt]
Method &$F$ &$f$ &Param (M) &FLOPs (G) &MPJPE $\downarrow$ \\
\midrule [0.5pt]

MixSTE \cite{mixste} &81 &81 &33.70 &92.42 &42.7 \\

MixSTE \cite{mixste} &147 &147 &33.73 &167.72 &41.8 \\

MixSTE \cite{mixste} &243 &243 &33.78 &277.25 &40.9 \\

\midrule [0.5pt]

\blue{HoT} w. MixSTE &243 &81 &35.00 &167.52 &{41.0} \\

\blue{TPC} w. MixSTE &243 &81 &33.78 &161.73 &\textbf{\red{40.4}} \\

\bottomrule [1pt]
\end{tabular}
\label{table:FLOPs_supp}
\VspaceSUPP
\end{table}

\section{Additional Quantitative Results}
\label{sec:Quantitative Results}
\noindent \textbf{Training Memory Cost and Training Time.}
To demonstrate the superiority of deploying our boosted VPTs on resource-limited devices, 
we report the training GPU memory cost and training time per epoch in Table~\ref{table:memory_supp} (directly using their training settings). 
Besides, we report the results of our method using the default settings, \textit{i.e.}, \{$F{=}351$, $n{=}1$, $f{=}117$\} for MHFormer, \{$F{=}243$, $n{=}3$, $f{=}81$\} for MixSTE, and \{$F{=}243$, $n{=}1$, $f{=}81$\} for MotionBERT. 
The results show that our method significantly reduces the GPU memory cost and training time while achieving superior results. 
For instance, HoT w. MotionBERT achieves a memory cost reduction of 43.0\% and a training time reduction of 47.5\% while maintaining the same performance. 

\noindent \textbf{Computation Complexity and Accuracy.}
In our main paper, we mainly report the results to show that our method can reduce FLOPs while achieving highly competitive or even better results (Tables ~\ref{table:pipeline}, \ref{table:block}, \ref{table:number}, and \ref{table:h36m}). 
Here, we compare our method with MixSTE using the same number of representative tokens and approximately the same number of FLOPs. 
To achieve this, we set the input frame number of the original MixSTE to $F {=} 81$ and $F {=} 147$, respectively. 
The results in Table~\ref{table:FLOPs_supp} show that our method obtains better results under both settings, further demonstrating the importance of large receptive fields and the effectiveness of our method.

\section{Additional Ablation Study}
\label{sec:Ablation Study}
\noindent \textbf{Number of Recovered Tokens.}
In Table~\ref{table:recovered_supp}, we conduct the ablation study on the number of recovered tokens ($f^{\prime}$) under \textit{seq2frame} pipeline. 
Since $f^{\prime}$ differs from the input frames, we evaluate the performance under \textit{seq2frame} pipeline, which selects the 3D pose of the center frame as the final estimation. 
The results show that reducing $f^{\prime}$ slightly decreases the number of parameters, but the performance remains almost unchanged. 
Therefore, we choose $f^{\prime} {=} 243$, which is more efficient and can be evaluated under \textit{seq2seq}. 

\begin{table}[t]
\footnotesize
\centering
\caption
{ 
Ablation study on the number of recovered tokens ($f^{\prime}$) under \textit{seq2frame} pipeline.
}  
\setlength{\tabcolsep}{2.10mm}
\begin{tabular}{l|cc|c}
\toprule [1pt]
Method &Param (M) &FLOPs (G) &MPJPE $\downarrow$ \\
\midrule [0.5pt]

MixSTE \cite{mixste} &33.78 &277.25 &40.9 \\

\midrule [0.5pt]

\blue{HoT} w. MixSTE ($f^{\prime} {=} 9$) &34.88 &163.33 &{40.9} \\

\blue{HoT} w. MixSTE ($f^{\prime} {=} 27$) &34.89 &163.66 &{40.7} \\

\blue{HoT} w. MixSTE ($f^{\prime} {=} 81$) &34.92 &164.62 &{40.9} \\

\blue{HoT} w. MixSTE ($f^{\prime} {=} 243$) &35.00 &167.52 &{40.9} \\

\bottomrule [1pt]
\end{tabular}
\label{table:recovered_supp}
\VspaceSUPP
\end{table}

\begin{table*}[t]
\footnotesize
\centering
\caption
{ 
Ablation study on the block index of representative tokens ($n$) under \textit{seq2seq} pipeline. 
}  
\setlength{\tabcolsep}{3.43mm}
\begin{tabular}{l|cllll|c}
\toprule [1pt]
Method &Param (M) &FLOPs (G) &FPS &GPU Memory (G) &Training Time &MPJPE $\downarrow$ \\
\midrule [0.5pt]

MixSTE \cite{mixste} &33.78 &277.25 &10432 &\ 
\ 11.4 &\ \ 17.0 &40.9 \\

\midrule [0.5pt]

\blue{HoT} w. MixSTE, $n {=} 1$  &35.00  &121.31 (\textbf{\blue{$\downarrow$  56.3\%}}) &20374 (\textbf{\blue{$\uparrow$ 95.3\%}}) &\ 
\ 6.0 (\textbf{\blue{$\downarrow$ 47.4\%}}) &\ \ 7.8  (\textbf{\blue{$\downarrow$ 54.1\%}}) &41.8 \\

\blue{HoT} w. MixSTE, $n {=} 2$  &35.00  &144.42 (\textbf{\blue{$\downarrow$ 47.9\%}}) &17724 (\textbf{\blue{$\uparrow$ 69.9\%}}) &\ 
\ 6.8  (\textbf{\blue{$\downarrow$ 40.4\%}}) &\ \ 9.2 (\textbf{\blue{$\downarrow$ 45.9\%}})  &41.6 \\

\blue{HoT} w. MixSTE, $n{=}3$ &35.00 &167.52 (\textbf{\blue{$\downarrow$ 39.6\%}}) &15770 (\textbf{\blue{$\uparrow$ 51.2\%}}) &\ 
\ 7.6 (\textbf{\blue{$\downarrow$ 33.3\%}}) &10.5 (\textbf{\blue{$\downarrow$ 38.2\%}})  &\textbf{\red{41.0}} \\

\blue{HoT} w. MixSTE, $n {=} 4$  &35.00  &190.62 (\textbf{\blue{$\downarrow$ 31.2\%}}) &14094 (\textbf{\blue{$\uparrow$ 35.1\%}}) &\ 
\ 8.5 (\textbf{\blue{$\downarrow$ 25.4\%}}) &12.0 (\textbf{\blue{$\downarrow$ 29.4\%}})  &41.4 \\

\blue{HoT} w. MixSTE, $n{=}5$ &35.00 &213.72 (\textbf{\blue{$\downarrow$ 22.9\%}}) &12801 (\textbf{\blue{$\uparrow$ 22.7\%}}) &\ 
\ 9.3 (\textbf{\blue{$\downarrow$ 18.4\%}}) &13.2 (\textbf{\blue{$\downarrow$ 22.4\%}}) &41.7 \\

\blue{HoT} w. MixSTE, $n {=} 6$  &35.00  &236.82 (\textbf{\blue{$\downarrow$ 14.6\%}}) &11673 (\textbf{\blue{$\uparrow$ 11.9\%}}) &10.0 (\textbf{\blue{$\downarrow$ 12.3\%}}) &14.7 (\textbf{\blue{$\downarrow$ 13.5\%}})  &41.6 \\

\blue{HoT} w. MixSTE, $n{=}7$ &35.00 &259.93 (\textbf{\blue{$\downarrow$ 06.3\%}}) &10791 (\textbf{\blue{$\uparrow$ 03.4\%}}) &10.9 (\textbf{\blue{$\downarrow$ 04.4\%}}) &16.0 (\textbf{\blue{$\downarrow$ 05.9\%}}) &41.5 \\

\bottomrule [1pt]
\end{tabular}
\label{table:block_supp}
\end{table*}

\begin{table*}[t]
\footnotesize
\centering
\caption
{ 
Ablation study on the number of representative tokens ($f$) under \textit{seq2frame} pipeline. 
Here, $^{*}$ denotes the result without re-training.
}  
\setlength{\tabcolsep}{2.38mm}
\begin{tabular}{l|cllll|cc}
\toprule [1pt]
Method &Param (M) &FLOPs (G) &FPS &GPU Memory (G) &Training Time &MPJPE$^{*}$ &MPJPE $\downarrow$ \\
\midrule [0.5pt]

MixSTE \cite{mixste} &33.78 &277.25 &43 &11.4 &\ \ 17.0 &40.7 &40.7 \\
\midrule [0.5pt]

\blue{TPC} w. MixSTE, $f {=} 9$ &33.78 &110.39 (\textbf{\blue{$\downarrow$ 60.2\%}}) &89 (\textbf{\blue{$\uparrow$ 107.0\%}}) &5.8 (\textbf{\blue{$\downarrow$ 49.1\%}}) &\ \ 7.2 (\textbf{\blue{$\downarrow$ 57.6\%}})  &44.1 &41.5 \\

\blue{TPC} w. MixSTE, $f{=}16$ &33.78 &115.38 (\textbf{\blue{$\downarrow$ 58.4\%}}) &88 (\textbf{\blue{$\uparrow$ 104.7\%}}) &5.8 (\textbf{\blue{$\downarrow$ 49.1\%}})  &\ \ 7.5 (\textbf{\blue{$\downarrow$ 55.9\%}}) &42.7 &41.0 \\

\blue{TPC} w. MixSTE, $f{=}27$ &33.78 &123.23 (\textbf{\blue{$\downarrow$ 55.6\%}}) &84 (\textbf{\blue{$\uparrow$ 95.3\%}}) &6.1 (\textbf{\blue{$\downarrow$ 46.5\%}}) &\ \ 7.9 (\textbf{\blue{$\downarrow$ 53.5\%}}) &41.8 &40.5 \\

\blue{TPC} w. MixSTE, $f{=}61$ &33.78 &147.47 (\textbf{\blue{$\downarrow$ 46.8\%}}) &75 (\textbf{\blue{$\uparrow$ 74.4\%}}) &7.0 (\textbf{\blue{$\downarrow$ 38.6\%}})  &\ \ 9.3 (\textbf{\blue{$\downarrow$ 45.3\%}})  &41.2 &40.5 \\

\blue{TPC} w. MixSTE, $f{=}81$ &33.78 &161.73 (\textbf{\blue{$\downarrow$ 41.7\%}}) &68 (\textbf{\blue{$\uparrow$ 58.1\%}}) &7.3 (\textbf{\blue{$\downarrow$ 36.0\%}}) &10.1 (\textbf{\blue{$\downarrow$ 40.6\%}})  &41.1 &40.4 \\

\blue{TPC} w. MixSTE, $f{=}121$ &33.78 &190.26 (\textbf{\blue{$\downarrow$ 31.4\%}}) &62 (\textbf{\blue{$\uparrow$ 44.2\%}}) &8.3 (\textbf{\blue{$\downarrow$ 27.2\%}})  &11.7 (\textbf{\blue{$\downarrow$ 31.2\%}})  &40.9 &40.4 \\

\blue{TPC} w. MixSTE, $f{=}135$ &33.78 &200.24 (\textbf{\blue{$\downarrow$ 27.8\%}}) &58 (\textbf{\blue{$\uparrow$ 34.9\%}}) &8.5 (\textbf{\blue{$\downarrow$ 25.4\%}})  &12.4 (\textbf{\blue{$\downarrow$ 27.1\%}})  &\textbf{\red{40.9}} &\textbf{\red{40.2}} \\

\bottomrule [1pt]
\end{tabular}
\label{table:number_supp}
\end{table*}

\begin{figure*}[tb]
\centering
\includegraphics[width=1.00\linewidth]{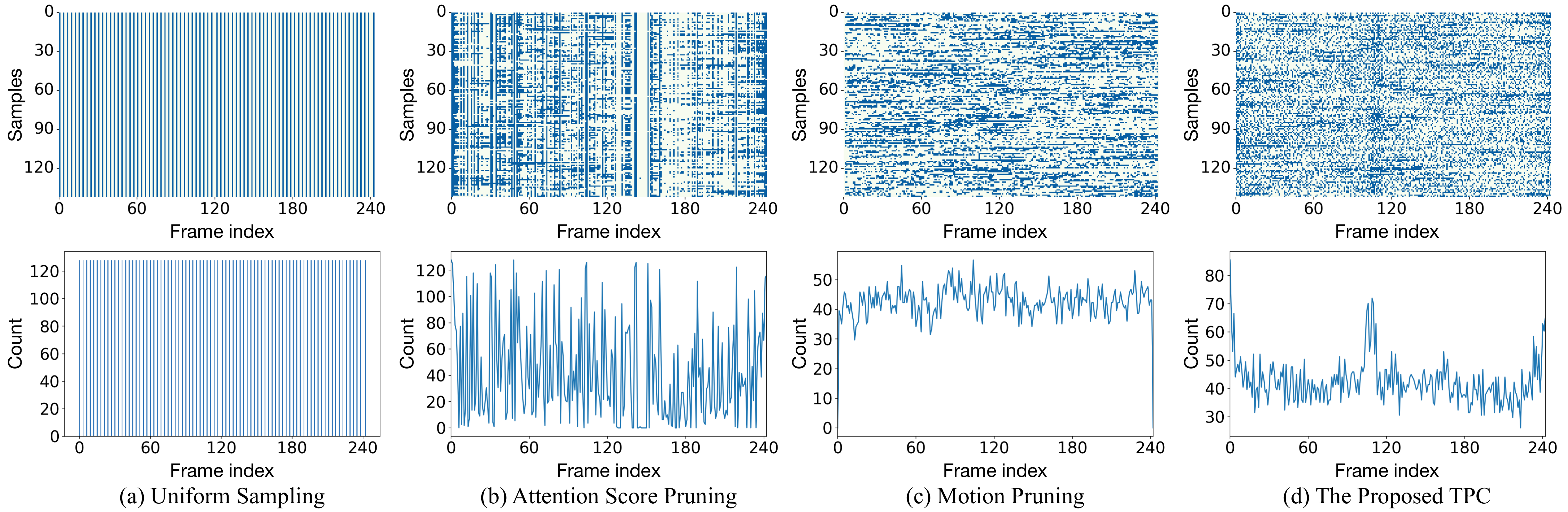}
\caption
{
Statistics visualization of selected tokens for different token pruning strategies. 
\textbf{Top}: Frame indexes of selected tokens for some samples (140 samples) of video sequences (243 frames). 
Blue points represent selected tokens and white points represent pruned tokens. 
\textbf{Bottom}: Frequency count of frame indexes of selected tokens for these samples. 
}
\label{fig:index_supp}
\end{figure*}

\begin{figure*}[tb]
\centering
\includegraphics[width=1.00\linewidth]{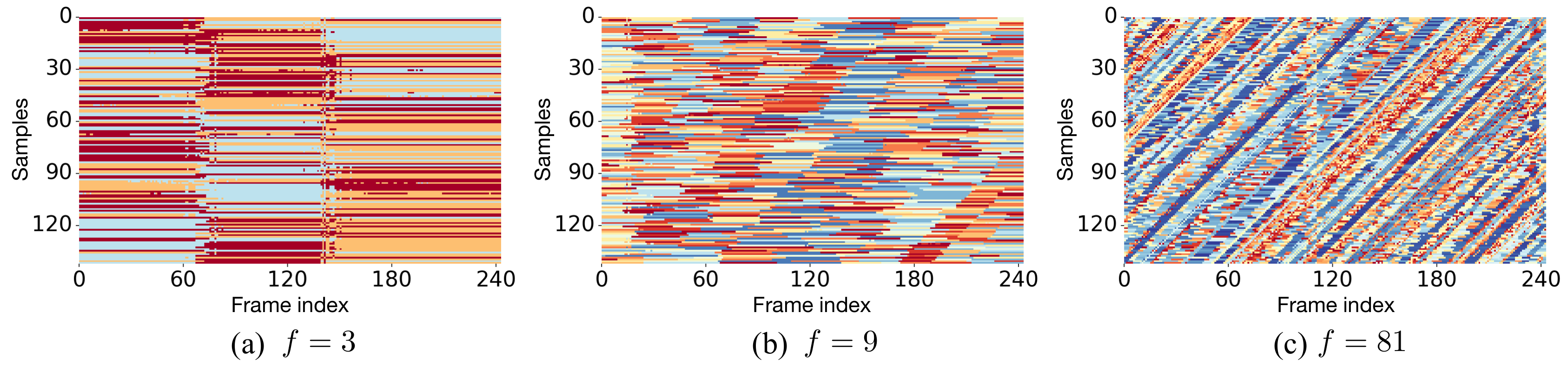}
\caption
{
    Visualization of cluster groups for the different numbers of representative tokens $f$. 
    In each row, points of the same color represent the same cluster group. 
}
\label{fig:cluster_supp}
\end{figure*}

\noindent \textbf{Hyperparameters ($n$ and $f$).}
In Tables~\ref{table:block} and \ref{table:number}, we conduct ablation studies on the block index of representative tokens ($n$) under the \textit{seq2frame} pipeline and on the number of representative tokens ($f$) under the \textit{seq2seq} pipeline, respectively. 
To systematically explore the hyperparameters, we further conduct the ablation studies on $n$ under the \textit{seq2seq} pipeline (Table~\ref{table:block_supp}) and on $f$ under the \textit{seq2frame} pipeline (Table~\ref{table:number_supp}). 
It shows that we can flexibly adjust the values of $n$ and $f$ to achieve a speed-accuracy trade-off that meets the specific demands of real-world applications. 

\section{Additional Visualization Results}
\label{sec:Visualization Results}
\noindent \textbf{Selected Tokens.}
In Figure~\ref{fig:index}, we provide statistics visualization of selected tokens by taking some samples of consecutive video frames as input with a temporal interval of 1 between neighboring samples. 
For more comprehensive observation, we further statistically visualize selected tokens of different token pruning strategies using random samples (temporal interval is set to 243), \textit{i.e.}, the neighboring samples have no overlapping frames. 
The frame indexes and frequency count of frame indexes of selected tokens are shown at the top and bottom of Figure~\ref{fig:index_supp}. 
The visualization figure of motion pruning (Figure~\ref{fig:index_supp} (c)) shows the most significant changes compared to Figure~\ref{fig:index}. 
The reason for this is that the random samples do not contain consecutive motion information.
Interestingly, the visualization figures of frame indexes between TPC and motion pruning are somewhat similar but our TPC selects more tokens for the center frame. 
Besides, the performance of motion pruning is much worse than our TPC due to noise frames (see Table~\ref{table:pruning}). 

\noindent \textbf{Cluster Groups.}
The visualization in Figure~\ref{fig:cluster_supp} depicts cluster groups corresponding to varying numbers of representative tokens ($f$). 
We observe that the cluster primarily groups neighboring tokens into the same group, as the nearby poses are similar. 
Moreover, it also groups some tokens that are relatively distant from each other into the same group based on their feature similarity. 

\begin{figure*}[tb]
\centering
\includegraphics[width=0.8\linewidth]{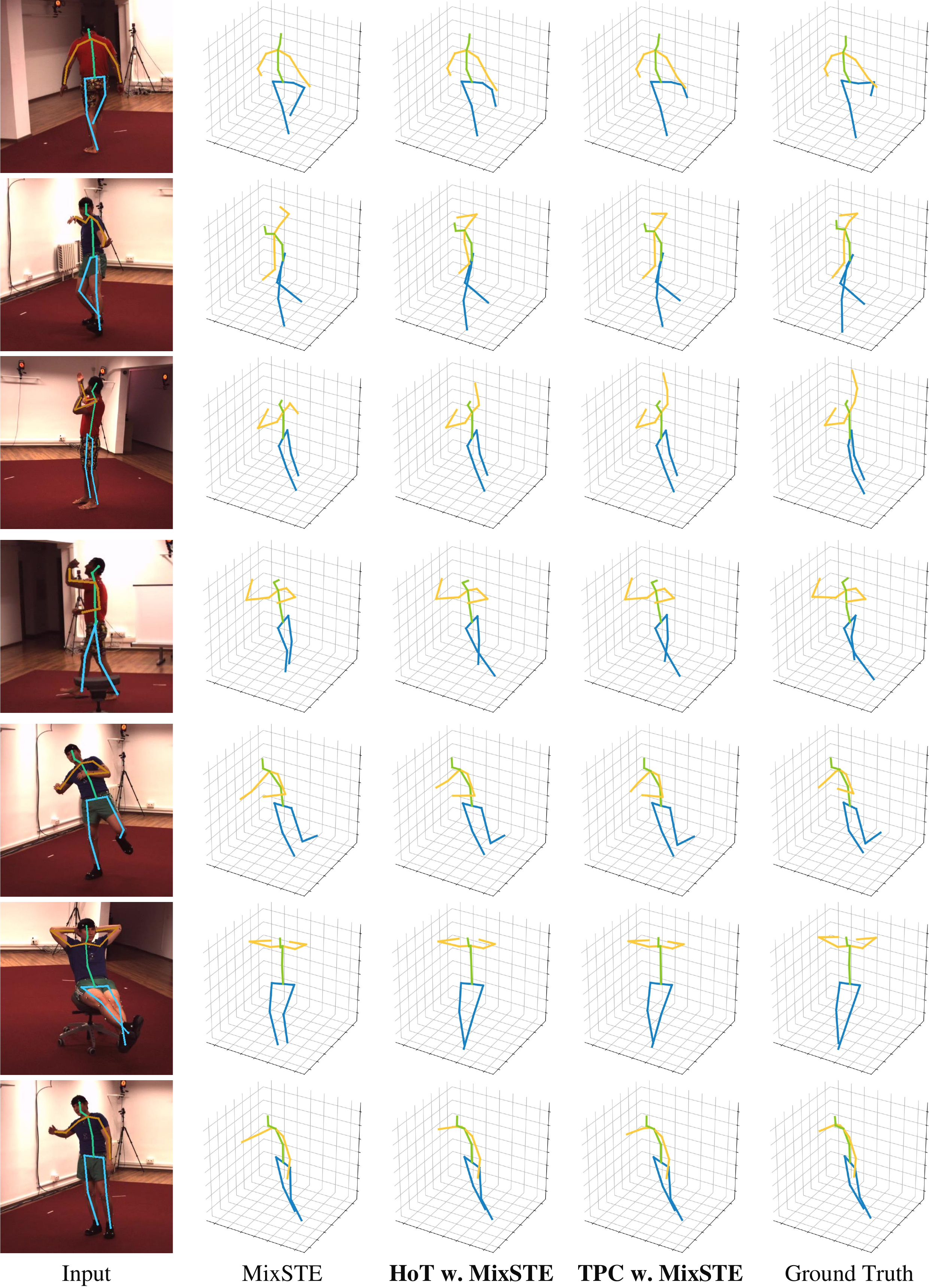}
\caption
{
    Qualitative comparison among the previous state-of-the-art method (MixSTE \cite{mixste}), our HoT w. MixSTE, and our TPC w. MixSTE on Human3.6M dataset. 
}
\label{fig:dataset_supp}
\end{figure*}

\begin{figure*}[tb]
\centering
\includegraphics[width=1.0\linewidth]{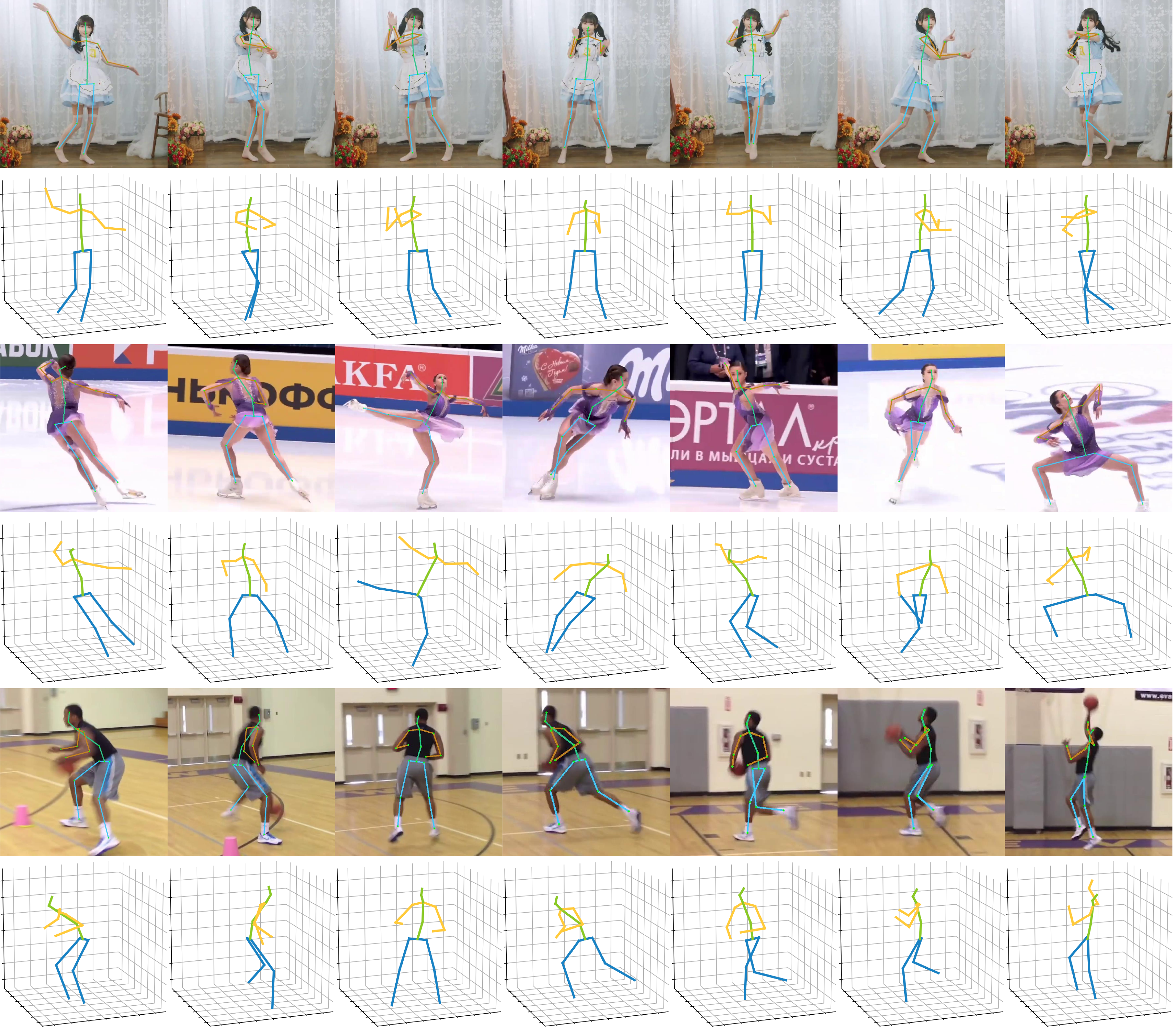}
\caption
{
    Qualitative results of our method on challenging in-the-wild videos. 
}
\label{fig:wild_supp}
\end{figure*}

\begin{figure*}[tb]
\centering
\includegraphics[width=0.93\linewidth]{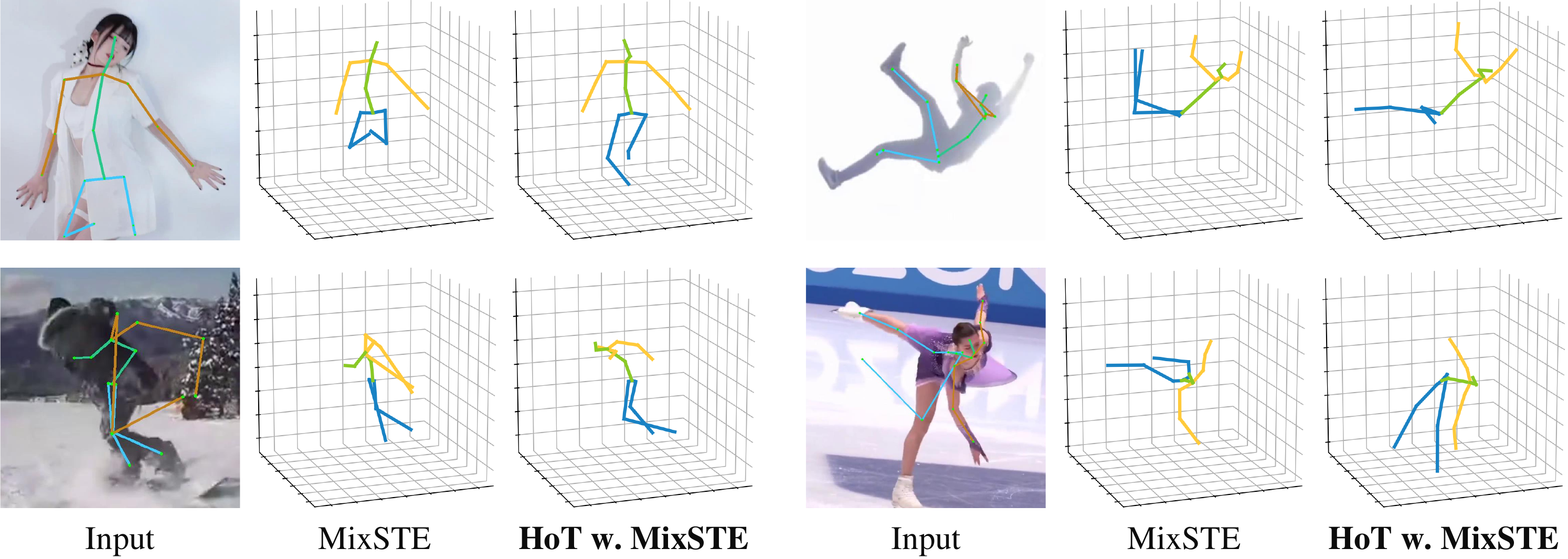}
\caption
{
    Failure cases in challenging scenarios. 
}
\label{fig:fail_supp}
\end{figure*}

\begin{figure*}[tb]
\centering
\includegraphics[width=1.0\linewidth]{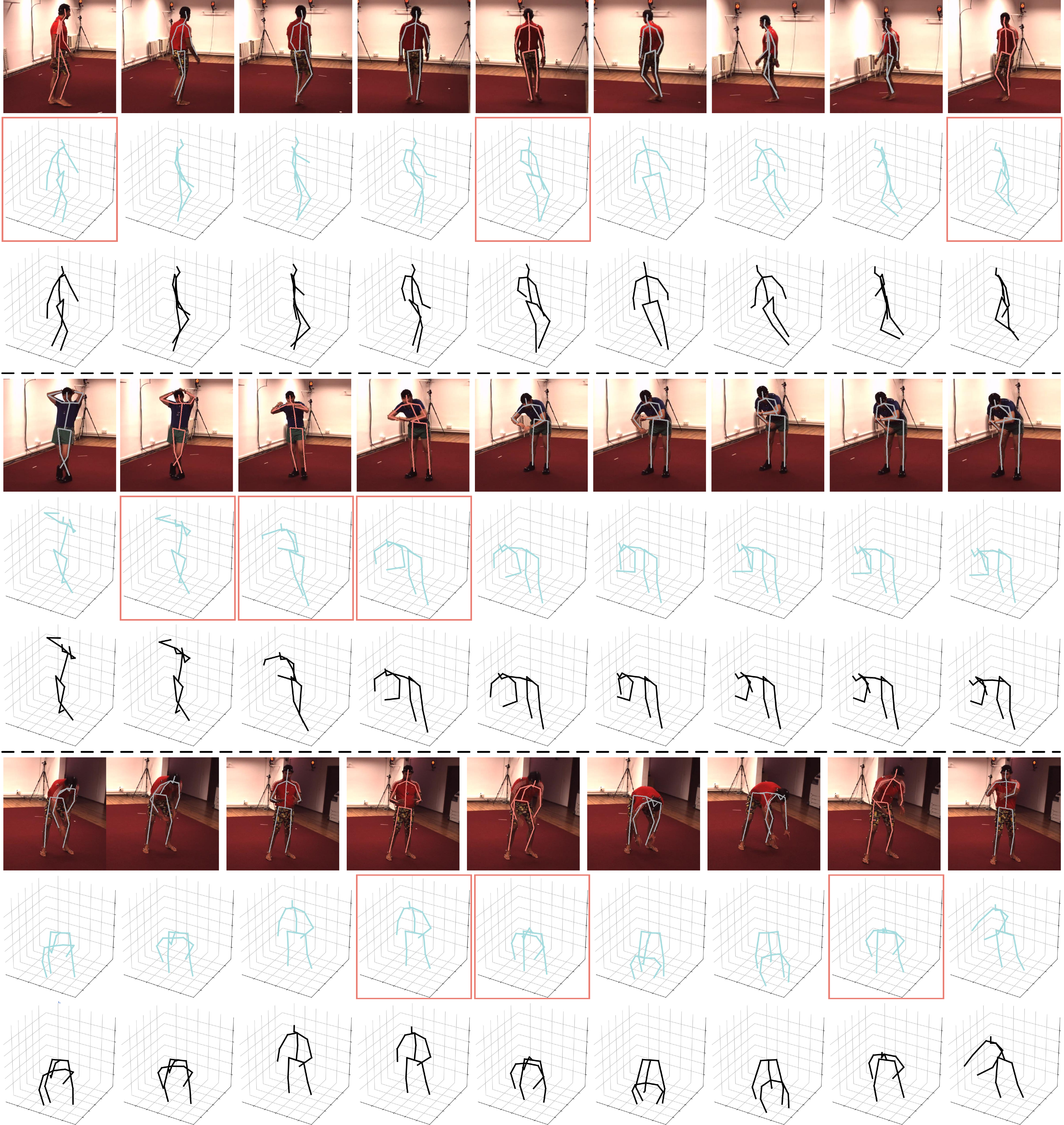}
\caption
{
    Visualization of input images, estimated 3D poses (cyan), and ground truth 3D poses (black) from three video sequences. 
    The 2D poses of selected frames are colored in red, and the 2D poses of pruned frames are colored in gray. 
    The 3D poses of selected frames are highlighted with red rectangular boxes. 
}
\label{fig:recovey_supp}
\end{figure*}

\noindent \textbf{3D Pose Reconstruction.}
Figure~\ref{fig:dataset_supp} presents the qualitative comparison among the proposed HoT w. MixSTE and TPC w. MixSTE, and MixSTE \cite{mixste} on Human3.6M dataset. 
Furthermore, Figure~\ref{fig:wild_supp} shows the qualitative results on challenging in-the-wild videos. 
These results confirm the ability of our method to produce accurate 3D pose estimations. 
However, in challenging scenarios, there are some failure cases where our method cannot accurately estimate 3D human poses due to factors such as partial body visibility, rare poses, and significant errors in the 2D detector (Figure~\ref{fig:fail_supp}). 
We also provide visualizations of recovering 3D human poses in Figure~\ref{fig:recovey_supp}, which illustrate that our method can predict realistic 3D human poses of the entire sequence, thereby further demonstrating the effectiveness of TRA. 

\end{document}